\documentclass[aps,prd,superscriptaddress,twoside,onecolumn,nofootinbib,showpacs,notitlepage]{revtex4-1}
\usepackage{amsmath,amssymb}
\usepackage{mathtools}
\usepackage[usenames]{color}
\usepackage{braket}
\usepackage{graphicx,bm}
\usepackage{slashed}
\usepackage{ulem}
\usepackage{slashbox}
\usepackage{algorithm} 
\usepackage{algpseudocode}
\usepackage{longtable}
\usepackage{adjustbox}
\usepackage{dsfont}
\usepackage{mhchem}
\usepackage[usenames]{color}
\renewcommand\sout{\bgroup \color{red} \ULdepth=-.5ex \ULset}

\newcommand{\beginsupplement}{%
        \setcounter{table}{0}
        \renewcommand{\thetable}{S\arabic{table}}%
        \setcounter{figure}{0}
        \renewcommand{\thefigure}{S\arabic{figure}}%
        \newcounter{SIfig}
        \renewcommand{\theSIfig}{S\arabic{SIfig}}}

\newcommand{\comment}[1]{}
\begin{document}

\comment{\title{\boldmath Evolution-based discovery of governing equations using deep neural networks}}
\title{\boldmath Neural-Network-Directed Genetic Programmer for Discovery of Governing Equations}

\author{Shahab Razavi}%
\email{shahab.razavi@vumc.org}
\affiliation{Division of Genetic Medicine, Department of Medicine, Vanderbilt University Medical Center, 
	Nashville, TN 37232, USA}

\author{Eric R. Gamazon}%
\email{ericgamazon@gmail.com}
\affiliation{Vanderbilt Genetics Institute, Vanderbilt University Medical Center, Nashville, TN 37232, USA}
\affiliation{Data Science Institute, Vanderbilt University Medical Center, Nashville, TN 37232, USA}
\affiliation{Clare Hall, University of Cambridge, Cambridge CB3 9AL, UK}

\date{\today}

\begin{abstract}
We develop a symbolic regression framework for extracting the governing mathematical expressions from observed data. The evolutionary approach, faiGP, is designed to leverage the properties of a function algebra that have been encoded into a grammar, providing a theoretical guarantee of universal approximation and a way to minimize bloat. In this framework, the choice of operators of the grammar may be informed by a physical theory or symmetry considerations. Since there is currently no theory that can derive the `constants of nature', an empirical investigation on extracting these coefficients from an evolutionary process is of methodological interest. We quantify the impact of different types of regularizers, including a diversity metric adapted from studies of the transcriptome and a complexity measure, on the performance of the framework. Our implementation, which leverages neural networks and a genetic programmer, generates non-trivial symbolically equivalent expressions (``Ramanujan expressions'') or approximations with potentially interesting numerical applications. To illustrate the framework, a model of ligand-receptor binding kinetics, including an account of gene regulation by transcription factors, and a model of the regulatory range of the cistrome from omics data are presented. This study has important implications on the development of data-driven methodologies for the discovery of governing equations in experimental data derived from new sensing systems and high-throughput screening technologies.
\end{abstract}


\maketitle

\section*{Introduction}\label{sec:Intro} 
Discovering a mathematical expression that accurately describes the governing equations from experimental data is of fundamental importance across all scientific fields. Traditionally, these expressions are formulated based on first principles such as conservation laws, known underlying symmetries, and other simplifying physical assumptions. With the growth of experimental data collection from new sensing systems and  high-throughput screening technologies in such diverse fields as physics, cosmology, molecular modeling, and genomics, the core task of data analysis, i.e., finding patterns and their mathematical formulations, presents new algorithmic and computational challenges. The development of new data-driven model-discovery methodologies has naturally become a main focus of recent research efforts \cite{Champion22445,Rudye1602614,MASLYAEV2019367, Brunton3932}. 

Symbolic regression (SR) aims to identify relationships in the data by searching the space of tractable mathematical expressions for the best-fit model. Specifically, given an observed sample $(X, y)$, with $X = \{\bm{x}_i\}_{i=1}^N$ and $y = \{y_i\}_{i=1}^N$, where each point $\bm{x}_i \in \mathbb{R}^d$ and $y_i \in \mathbb{R}$, SR seeks to find a tractable expression $\tau: K \subset \mathbb{R}^d \rightarrow \mathbb{R}$ on the domain $K$ which can accurately describe the unobserved generative function $f$ with $f(\bm{x}_i) = y_i$ up to an arbitrarily small error, $|\tau(\bm{x}_i) - f(\bm{x}_i)| \leq \epsilon$ for $1 \leq i \leq N$, over $X$. Exact recovery of $f$ or an equivalent mathematical expression is desired; in practice, an approximating function at a pre-specified level of accuracy can facilitate important applications. $\tau$ can be represented as a sequence consisting of unary operators (e.g., sine), binary operators (e.g., $+$), and operands (e.g., variables and constants). Despite the notable progress in this area, the general symbolic regression problem is likely to be NP-hard \cite{HYAFIL197615}, rendering a general solution out of reach. Nevertheless, equations of practical interest in the sciences that describe natural laws can generally be expressed in a simple form. Consequently, in practice, several proposed algorithms aim to reduce the exponentially large search space of functions, making the search for a tractable expression manageable. Feasibility is achieved by exploring a targeted function search space, generally defined by a preselected set of operators and operands. Furthermore, to ensure the size of the search space does not grow exponentially with the length of the expression, a cap on the expression's length is set. The expression is then evaluated using a fitness score. 

Exhaustively searching through the set of possible strings of symbols (encoding the mathematical expression), i.e., the space of computationally feasible expressions, and evaluating their fitness score is the most resource intensive and computationally demanding element of existing algorithms. An early effort to address this challenge is Genetic Programming (GP) \cite{Koza1994}, in which an algorithm heuristically searches over a large space of expressions. Searching this space uniformly and merely on the basis of goodness of the fit results in slow and unreliable convergence. Liskowski et al. \cite{NeuroGuidedGP} exploited the idea that additional domain knowledge can accelerate the search. The domain knowledge can be acquired using DeepCoder, a deep neural network, proposed in \cite{balog2017deepcoder}. DeepCoder helps to guide GP to search the space more effectively using a prior probability distribution of instructions in an expression. Although the achieved boost in performance was not statistically significant, the results were promising. Similarly, Haeri et al. in Statistical Genetic Programming (SGP) \cite{AMIRHAERI2017447} used statistical information, including variance, mean, and correlation coefficient, to improve GP. Their results showed an improvement in the evolutionary rate, the accuracy of the solutions, generalizability, and a decrease in the rate of code growth. However SGP is not guaranteed to find the optimal solution.

Searching the solution space of SR, the discrete space of computationally feasible expressions, poses an inherent difficulty for any algorithm. The discreteness means the correlation between changes in the expression and changes in the evaluated fitness of said expression can not be simply predicted. In other words, any local information in the search space is destroyed and any small change in the expression could result in a large shift in the fitness. Some attempts have been made to modify GP to work in a pseudo-continuous manner. For example, Anjum et al. \cite{anjum2019novel} explores the idea of using recurrent neural networks (RNN) to construct the gene strings. Then a powerful continuous evolutionary algorithm optimizes the weights of the network. The RNN mapping smoothens the sharp fitness landscape. This method has been shown to have the potential to improve accuracy and efficiency on several well-known SR problems, but the scope is limited.

In GrammarVAE, Kusner et al. \cite{kusner2017grammar} developed a variational autoencoder that can generate parse trees, representing discrete objects using a context-free grammar. The authors demonstrated that their model can be used for symbolic regression. However, despite decoding into parse trees, the method could not always produce syntactically valid expressions, and it struggled to exactly recover benchmark expressions. 

Udrescu et al. \cite{Udrescueaay2631} proposed a recursive multidimensional symbolic regression algorithm that uses neural networks to discover hidden properties such as symmetry or separability in the dataset to break the problem into simpler sub-problems with fewer variables. These sub-problems then can be solved using simple techniques like polynomial fitting. In a later publication \cite{udrescu2020ai}, the authors further improved on this approach by using Pareto-frontier to prune the search space and taking advantage of statistical hypothesis testing for more robustness.

Petersen et al. \cite{petersen2021deep} presented an approach that involves representing mathematical expressions as sequences. Using a recurrent neural network to emit a distribution over tractable mathematical expressions, they developed an autoregressive model that generates expressions under a pre-specified set of constraints. They used a risk-seeking policy gradient to train the model to generate better-fitting expressions. 

Here we present an alternative to standard GP, which we call Function-Algebra-Informed Genetic Programming (faiGP), a new GP framework, with a core grammar built on the properties of a function algebra. The approach uses the power of variational autoencoders \cite{kingma2014autoencoding, rezende2014stochastic} with two convolutional layers (ConVAE) to perform dimensionality reduction and a Bayesian classifier to generate a prior distribution over operators and operands. This probability distribution guides the faiGP both in generating its initial population and in the evolutionary process. In contrast to standard GP, faiGP is informed by a function algebra with a new data structure at its core to represent programs during the evolutionary process. GP uses binary trees to represents programs; binary trees have the advantage of being very efficient to program due to its simplicity. This simplicity, nevertheless, has been linked to many of the shortcomings of GP such as bloat \cite{10.1007/3-540-45493-4_48}. To remedy some of the causes of the bloat, we propose a grammar for generating programs alongside a new data structure that complies with the grammar rules. As a result, our approach also updates the evolutionary process such that the rules of the grammar will not be violated during the evolution. 

\section*{Methods}\label{sec:Methods}
\subsection*{Background}\label{sec:Background}

We begin by reviewing the theoretical foundations of the core components of our methodology. Our SR approach integrates dimensionality reduction (to reduce the search space), posterior inference (to guide the initial step), and evolution of programs (to find the best model in terms of fitness and complexity).
\par\null\par
\noindent \textbf{Variational autoencoder.}\label{subsec:VAE} The variational autoencoder (VAE), introduced by Kingma and Welling \cite{kingma2014autoencoding}, uses stochastic variational inference to map the input data $X = \{ \bm{x}_i \in \mathbb{R}^d \}_{i=1}^N$ onto a latent space. The data is assumed to be generated randomly, involving a continuous random variable $\bm{z}$ with a prior distribution $p_{\theta^*}(\bm{z})$. A value $\bm{x}_i$ is then generated from some conditional distribution $p_{\theta^*}(\bm{x}|\bm{z})$. The true parameters $\theta^*$ and the values of the latent variable $\bm{z}$ are hidden. We assume that the prior $p_{\theta^*}(\bm{z})$ and the likelihood $p_{\theta^*}(\bm{x}|\bm{z})$ belong to parametric families of distributions $p_{\theta}(\bm{z})$ and $p_{\theta}(\bm{x}|\bm{z})$. In general, the integral of the marginal likelihood $p_\theta(\bm{x}) = \int p_\theta(\bm{z}) p_\theta(\bm{x}|\bm{z})d\bm{z}$ is intractable, hence the posterior density $p_{\theta}(\bm{z}|\bm{x}) = p_\theta(\bm{x}|\bm{z})p_\theta(\bm{z})/p_\theta(\bm{x})$ is intractable.

In order to overcome this problem, VAE approximates the true (intractable) posterior distribution $p_{\theta}(\bm{z}|\bm{x})$ with a recognition model $q_\phi(\bm{z}|\bm{x})$, aka a probabilistic \textit{encoder}\comment{, and assumes a Gaussian distribution, $p(z)$, over the  possible values of the latent variable $z$}. The likelihood $p_{\theta}(\bm{x}|\bm{z})$ is referred to as a probabilistic \textit{decoder}. The parameters $\phi$ and $\theta$ (representing the weights and biases of a neural network) are jointly learned by maximizing the evidence lower bound (ELBO):

\begin{align}
	{\cal L} (\theta, \phi ; \bm{x}) &= \int \log(\frac{p_\theta(\bm{x}, \bm{z})} {q_\phi(\bm{z}|\bm{x})}) q_\phi(\bm{z}|\bm{x})d\bm{z} \nonumber \\
	&= \mathbb{E}_{q_\phi(\bm{z}|\bm{x})} \left[- \log q_\phi(\bm{z}|\bm{x}) + \log p_\theta(\bm{x}, \bm{z})  \right] \nonumber \\ 
	&= \mathbb{E}_{q_\phi(\bm{z}|\bm{x})} \left[- \log q_\phi(\bm{z}|\bm{x}) + \log p_\theta(\bm{x}|\bm{z}) \right. \nonumber \\
	& \hspace{0.5cm} \left. + \log p_\theta(\bm{z})  \right]
	\label{eq:ELBO}
\end{align}

\noindent where $\log p_\theta(\bm{x}) \ge {\cal L} (\theta, \phi ; \bm{x})$.

\par\null\par
\noindent \textbf{Bayesian multilabel classifier.}\label{subsec:BMC} Consider a dataset ${\cal D}$ consisting of $N$ tuples $(\bm{x}_i, \bm{y}_i)_{i=1}^N$ where $\{ \bm{x}_i \in \mathbb{R}^d \}_{i=1}^N$ and $\bm{y}_i \in \{0, 1\}^C$. $\bm{y}_i$ denotes a label vector, of size $C$, the number of classes. In the multi-label setting, $y_i^c = 1$ if example $\bm{x}_i$ belongs to class $c$, and $0$ otherwise. The goal is to tune the parameters $\bm{w}$ of a model $p(\bm{y}|\bm{x}, \bm{w})$ such that it predicts $\bm{y}$ given $\bm{x}$ and $\bm{w}$. Starting with some initial belief over the parameters $\bm{w}$ in the form of a prior distribution $p(\bm{w})$, Bayesian inference updates the prior distribution into a posterior distribution $p(\bm{w}|{\cal D}) = p({\cal D}|\bm{w}) p(\bm{w}) / p({\cal D})$. Once again, the right hand side usually involves computation of an intractable multidimensional integral. Using variational inference, we approximate $p(\bm{w}|{\cal D})$ by a parametric distribution $q_\phi(\bm{w})$. The quality of this approximation is measured by the Kullback-Leibler (KL) divergence $D_{KL}(q_\phi(\bm{w})||p(\bm{w}|{\cal D}))$. The divergence between the true posterior distribution and our approximated posterior is minimized in practice by maximizing the variational lower bound ${\cal L}(\phi)$:

\begin{equation}
    {\cal L}(\phi) = {\cal L}_{\cal D}(\phi) - D_{KL}(q_\phi(\bm{w})||p(\bm{w}))
\end{equation}
where
\begin{equation}
    {\cal L}_{\cal D}(\phi) = \sum_{(\bm{x}, \bm{y}) \in {\cal D}} \mathbb{E}_{q_\phi (\bm{w})} \left[ \log p(\bm{y}|\bm{x},\bm{w}) \right] \
    \label{eq:BMC_likelihood}
\end{equation}

\par\null\par
\noindent \textbf{Genetic programming.}\label{subsec:GP} The basic idea of genetic programming (GP) \cite{Koza1994} for SR is to evolve generations of programs, sequences of operators and operands, using a series of heuristic operations analogous to biological evolution. In the first generation, the programs are generated randomly, and each subsequent generation is produced according to the evolutionary process. Koza, in the first application of GP \cite{Koza1994}, used tree structures to represent programs in each generation. In this representation, the internal nodes hold the operators while the variables and constants are located in the leaves. While the tree structure can be easily implemented, it is not the most efficient in controlling bloat. Besides the tree-based representation, many other representations have been proposed over the years. Perkins in \cite{ieee94:perkis} presented the idea of using stacks for GP. Linear Genetic Programming (LGP) represents programs in a population as a sequence of instructions from imperative programming language or machine language \cite{10.5555/280485,Brameier2001ACO,kinnear:nordin}. Cartesian Genetic Programming (CGP) was developed \cite{10.5555/2934046.2934074} as a generalization of a method that had been used in electronics to encode and evolve digital circuits \cite{Miller97designingelectronic}. Later, CGP became established as a new form of genetic programming \cite{10.1007/978-3-540-46239-2_9}. In CGP, computational structures such as mathematical equations or computer programs are encoded as directed graphs. The advantages of graphs over trees are that graphs can model diverse systems, nodes can have multiple uses, and edges can express relationships flexibly. Graphs have the versatility to represent a wide range of computational structures such as systems of equations, state-machines, neural networks, algorithms, and electronic circuits \cite{Miller2020}.

\subsection*{Function-Algebra-Informed Genetic Programming}\label{sec:faiGP}

We describe faiGP as an evolutionary approach to the exploration of the space of expressions, iteratively transforming a population of programs into a new generation through the application of genetic operators. The evolutionary process ``sees'' the properties of a function algebra that have been encoded into a grammar, reducing the possibility of bloat and providing a universal approximation theoretical guarantee.
\par\null\par

\noindent \textbf{Evolutionary process.} As in conventional GP, faiGP assumes a fitness function $\gamma: {\cal P} \rightarrow {\cal M}$, where ${\cal P}$ is a set of functions $f: K \subset \mathbb{R}^d \rightarrow \mathbb{R}$ and ${\cal M}$ is a partially ordered set. Here, we assume ${\cal M} = \mathbb{R}$. GP subjects, according to pre-defined probabilities, a subset ${\cal P}_0$ of ${\cal P}$ to an evolutionary process to find an optimal point for $\gamma$. Let $n_{p}$ and $n_{o}$ be the number of parents and of offsprings, respectively. The evolutionary process includes the following evolutionary operators: mutation $m:{\cal P}_0^{n_{o}} \times \Theta_{m}  \rightarrow {\cal P}_0^{n_{o}}$, crossover $c:{\cal P}_0^{n_{p}} \times \Theta_{c} \rightarrow {\cal P}_0^{n_{o}}$, hoist $h:{\cal P}_0^{n_{p}} \times \Theta_{h} \rightarrow {\cal P}_0^{n_{o}}$, subtree $t:{\cal P}_0^{n_{o}} \times \Theta_{t}  \rightarrow {\cal P}_0^{n_{o}}$, and selection $s:{\cal P}_0^{n_{o}} \times \Theta_{s}  \rightarrow {\cal P}_0^{n_{p}}$, where each $\Theta_{*}$ is the probability space for the corresponding operator and gives the evolutionary process from one generation to the next a stochastic nature. 

Every element in ${\cal P}_0$ has a representation that can be produced from a grammar $G$ (defined below). We refer to each such element in ${\cal P}_0$ a $G$-representable function.

\par\null\par
\noindent \textbf{Search space, bloat, and program representation.} While standard GP has been shown to be successful in finding solutions to a range of problems, it suffers from several crucial drawbacks, as Haeri et al. noted \cite{AMIRHAERI2017447}:

\begin{itemize}
    \item Code growth, commonly known as bloat \cite{10.5555/222025}, is the phenomenon in which the length of the expression gradually increases without any noticeable improvement in the fitness score \cite{10.5555/1595536.1595563}. Bloat results in two challenges for GP. Firstly, the computational cost of evolving longer expressions grows exponentially. Secondly, with the increase in length or complexity of the expressions, the generalizability of the solution may decrease.
    Several theories have been proposed to explain the cause of bloat. Some of the most promising attempts are the removal bias theory
    \cite{soule:1998:rbias}, replication accuracy theory \cite{Mcphee95accuratereplication}, nature of program search space theory \cite{Langdon97fitnesscauses, 10.5555/316573.317113}, and crossover bias theory \cite{10.1145/1276958.1277277, 10.1007/978-3-540-71605-1_18}.
    
    \item An intractably large search space is another challenge for GP. The huge search space is the result of having a variable expression length irrespective of the function search space size. In this large search space, not every solution is unique \cite{782609} and there are many \textit{mathematically} or \textit{functionally} equivalent solutions, i.e., redundancies, e.g. $x(x+1)$ and $x^2+x$. These redundant expressions form an equivalence class induced by an equivalence relation; most of its members are large, long-sequence solutions according to the \textit{nature of program search space theory} \cite{10.1145/1276958.1277277, 10.1007/978-3-540-71605-1_18}. Each class can be represented with one unique solution program; we chose the expanded program, the program with no trivial parentheses, with no nullifying components, e.g. $x^2$ as opposed to $x+x^2-x$, as the class representative.
    
    \item Some problems are inherently difficult for GP \cite{10.1023/A:1011504414730}. Daida et al. \cite{10.1007/3-540-45110-2_60}. experimentally showed that this inherent difficulty could be the result of inefficiencies in how GP searches the solution space. Specifically, they showed that GP can not easily produce program trees with “high depth and small size,” or trees with “shallow depth and large size.” 
    
\end{itemize}

To address the challenges mentioned above, we propose a new approach to genetic programming using a grammar built on the properties of a function algebra with sets (instead of trees) for program representation, with two additional expanded ``union'' operations (see equations (\ref{def:dotCup}) and (\ref{def:dotCupTimes})). Set operations and properties within our proposed grammar (equation (\ref{eq:grammar})) facilitate the search to identify or generate the unique program that serves as the equivalence class representative; utilizing such a representative program results in the reduction or elimination of bloat in many cases. For example $x+x^2-x$ and $x^2$ belong to the same equivalence class and are functionally equivalent expressions. Identifying and eliminating such equivalences is difficult in the tree representation of programs, and several approaches have been proposed to overcome this issue such as calculating the subtree mean and variance \cite{AMIRHAERI2017447}, or code editing \cite{koza94book}. Our algorithm identifies $x^2$ as the class representative and eliminates all other functionally equivalent programs via set equality. Effectively, our search space is ${\cal P}_0/\sim$, i.e., the set of equivalence classes under the relation $\sim$, defined by: $f~\sim~g$ if and only if $f(\bm{k}) = g(\bm{k})$ for every $\bm{k} \in K$.

The smallest program, in our implementation, is a 4-tuple: a constant coefficient, an element that specifies the operator, a set that encompasses the operands to which the operator will be applied, and lastly an integer that specifies the exponent of the operator (equation (\ref{eq:grammar})). Each projection operator $\pi_i$, for $i \in \{1,2,3,4\}$ with $\pi_i(a_1, a_2, a_3, a_4) = a_i$, may be composed with the evolutionary operators, allowing each component of the 4-tuple to evolve independently according to pre-defined probabilities. A major difference between our approach and all previous approaches is that our approach treats the exponent explicitly as a component of a program (along with the coefficient, the operator, and the operand). For example, if $(x+1)^5$ is the unknown symbolic expression, standard GP will most likely generate this expression by randomly \textit{multiplying} $x$ many times in order to search for each exponent of $x$ independently, making the search for such expressions more difficult. The difficulty increases as the exponent grows, as Daida et al. have shown \cite{10.1023/A:1011504414730}. Allowing the exponent to independently evolve gives the framework more direct control to explore the space of possible expressions which could results in accelerating the search.

\par\null\par
\noindent \textbf{Grammar.} We define the grammar $G=(\{S, S_+, S_\times, O, O_+, O_\times, U, F, C, P\}, \{prod, sum, \cup, \dot{\cup}, \dot{\cup}_\times, x, r, z\}, R, S)$ as a four-element tuple, with the production rules $R$ and a start symbol $S$. The first element of the tuple $G$ is the set of non-terminal variables. Each of these variables represents a different type of program. $S_+$ generates a program in which every multiplication is distributed over addition; $S_\times$ generates a program that is expressed as the product of all elements; $O$ represents an operand, i.e., a set of real numbers and/or independent variables; $O_+$ generates an operand in $S_+$; $O_\times$ generates an operand in $S_\times$; $U$ generates the smallest program or a program representing the composition of functions; $F$ represents an element of the library ${\cal F}$, a pre-defined set of available operators; $C$ represents the coefficient of an operator; and finally $P$ represents the exponent of an operator. The second element of $G$, disjoint from the first element, is the set of terminal symbols consisting of the multinary operators \textit{prod} and \textit{sum}, the union operators $\cup$, $\dot{\cup}$, and $\dot{\cup}_\times$ (see equations (\ref{def:dotCup}) and (\ref{def:dotCupTimes}) for definition), the variable $x$, a real number $r \in \mathbb{R}$, and an integer $z \in \mathbb{Z}$. 
\comment{The $\dot{\cup}$ operator in $U ~ \dot{\cup} ~ S$ acts on $U / C$ and $S / C$ where $U / C = \{O, T, P \}$ and adds the coefficients if $U / C \in S / C$ otherwise it returns the union of $S$ and $U$. Meanwhile, $\dot{\cup}_\times$ in $U ~ \dot{\cup}_\times ~ S_{\times}$ acts on $U / \{C, P\}$ and $S_\times / \{C, P\}$ and if $U / \{C, P\} \in S_\times / \{C, P\}$ then the coefficients are multiplied and exponents are added, otherwise $U / \{C\}$ is added to $S_\times / \{C\}$ and the coefficients are multiplied.} 
The production rules $R$ are defined by:

\begin{align}
    1.& ~S \rightarrow S_{+} \mid S_{\times} \mid U \nonumber \\
    2.& ~S_{+} \rightarrow \{(C, sum, O_{+}, P)\} \nonumber \\
    3.& ~S_{\times} \rightarrow \{(C, prod, O_{\times}, P)\} \nonumber \\
    4.& ~O_{+} \rightarrow U ~\dot{\cup}~ O_{+} \mid U ~\dot{\cup}~ S_{\times} \mid U \nonumber \\
    5.& ~O_{\times} \rightarrow U ~\dot{\cup}_\times~ O_{\times} \mid U \nonumber \\
    6.& ~U \rightarrow \{(C, F, O, P)\} \mid \{(C, F, S, P)\} \nonumber \\
    7.& ~F \rightarrow \sqrt{} \mid \cos \mid \sin \mid \log \mid \mathrm{af\!fine} \nonumber \\
    8.& ~O \rightarrow O \cup \{x\} \mid O \cup \{r\} \mid \{x\} \mid \{r\} \nonumber \\
    9.& ~C \rightarrow r \nonumber \\
    10.& ~P \rightarrow z \nonumber \\
    \label{eq:grammar}
\end{align}

\noindent Here, the vertical bar means \textit{or}. We define the union operators $\dot{\cup}$ and $\dot{\cup}_\times$ as follows:

\begin{align}
    \label{def:dotCup}
    U_1 ~\dot{\cup}~ U_2 =
        \begin{cases}
            \{\{(C_1, F_1, O_1, P_1)\},\{(C_2, F_2, O_2, P_2)\}\} ~~ \textrm{if} ~~ \{(F_1, O_1, P_1)\} \neq \{(F_2, O_2, P_2)\} \\
            \{(C_1+C_2, F_1, O_1, P_1)\} ~~ \textrm{otherwise}
        \end{cases}
\end{align}

\noindent and 

\begin{align}
    \label{def:dotCupTimes}
    U_1 ~\dot{\cup}_\times~ U_2 =
        \begin{cases}
            \{\{(C_1, F_1, O_1, P_1)\},\{(C_2, F_2, O_2, P_2)\}\} ~~ \textrm{if} ~~ \{(F_1, O_1)\} \neq \{(F_2, O_2)\} \\
            \{(C_1*C_2, F_1, O_1, P_1+P_2)\} ~~ \textrm{otherwise}
        \end{cases}{}
\end{align}

\noindent where $U_1 = \{(C_1, F_1, O_1, P_1)\}$ and $U_2 = \{(C_2, F_2, O_2, P_2)\}$. $\dot{\cup}$ and $\dot{\cup}_\times$ reduce to the regular union operator $\cup$ when one of the sets to be combined using the operator is of representation $O$:

\begin{align}
    U_1 ~\dot{\cup}~ O = U_1 ~\dot{\cup}_\times~ O = U_1 ~\cup~ O.
\end{align}

The grammar encodes an algebraic structure with the closure property on the operations of addition, multiplication (with associativity), and scalar multiplication. Thus, the grammar describes an associative algebra (an $\mathbb{R}$-algebra). We note that the grammar is also closed under composition of operators (Rule 6), generating an additional layer of structure. The library ${\cal F}$ is chosen (Rule 7) so as to be compatible with the closure property under composition, i.e., produce a syntactically valid and well-defined expression. (Henceforth, the operator $\sqrt{}$ refers to the square root of the absolute value; similarly, the operator $\log$ refers to the natural log of the absolute value, with $\log(0) \coloneqq 0$.) We do not put strict constraints on compositions of operators (i.e., on allowed descendants of an operator), as these compositions appear in many potentially useful guises (e.g., $-\log|\cos(x)|$ as the antiderivative $\int \tan(x) dx$, up to a constant). Rule 8 enables production of functions of several variables.

By design, faiGP generates and evolves programs in compliance with this grammar. The smallest program is a set with one 4-tuple generated through $S \rightarrow U \rightarrow \{(C, F, O, P)\}$, i.e., the consecutive application of Rule 1 and Rule 6. Here, $F$ is an operator, $C$ is the coefficient of $F$, $O$ is an operand on which $F$ acts, and $P$ is the exponent of $F$. 

\par\null\par
\noindent \textbf{Universal approximation property.} Suppose ${\cal P} = C(K, \mathbb{R})$, the algebra of real-valued continuous functions on a compact Hausdorff space $K \subset \mathbb{R}^d$. We note that any monomial $x_i^n$, $1 \leq i \leq d$ for a positive integer $n \in \mathbb{N}$ can be derived from our grammar $G$ via the following production rules:  $P \rightarrow n$ (Rule 10), $C \rightarrow 1$ (Rule 9), $O \rightarrow \{x_i\}$ (Rule 8),  $F \rightarrow \mathrm{af\!fine}$ (Rule 7), $U \rightarrow \{(C, F, O, P)\}$ (Rule 6). By the Stone–Weierstrass theorem, the intersection of the subalgebra ${\cal P}_0$ of $G$-representable functions and ${\cal P}$ is a dense subset of ${\cal P}$ under the supremum norm topology since ${\cal P}_0$ contains the set $\mathbb{R}[x_1,\ldots, x_d]$ of polynomial functions with real coefficients: $\mathbb{R}[x_1, \ldots, x_d] \subseteq {\cal P}_0 \cap {\cal P} \subseteq {\cal P}$. (Note that, given our choice of $\cal F$, since $\log(x) \in {\cal P}_0$, ${\cal P}_0$ contains functions with discontinuities.) Thus, given $f \in {\cal P}$, for every $\varepsilon > 0$, there is a $G$-representable function $f_{\mathrm{approx}}$ such that $||f(\bm{k}) - f_{\mathrm{approx}}(\bm{k})|| < \varepsilon$ for every $\bm{k} \in K$. Thus, the constraint of the grammar $G$ does not take us far from the original function $f$. We summarize these observations in the following:

\par\null\par
\noindent \textbf{Theorem 1}: Any $f \in C(K, \mathbb{R})$ can be uniformly approximated by some function $f_{\mathrm{approx}}$ in the subalgebra ${\cal P}_0$ of $G$-representable functions.

\par\null\par
\noindent The following observation connects the grammar to deep learning.

\par\null\par
\noindent \textbf{Theorem 2}:  A feedforward neural network with Rectified Linear Unit (ReLU) is the uniform limit of $G$-representable functions.
\par\null\par
\noindent Proof:  Let $f_\sigma$ be a feedforward neural network with the ReLU activation function $\sigma$. By Theorem 1, the absolute value (a continuous operator) of a continuous function $g$, $\left| g \right|$, can be uniformly approximated by a $G$-representable function. Thus, $\max(g, h) = \frac{ g + h }{2} + \frac{\left| g - h \right|}{2}$ is also the uniform limit of $G$-representable functions. In particular, so is $\max(0, x_i)$, $1 \leq i \leq m$. The result follows since $f_\sigma$ is the alternating composition of an affine map and ReLU:

\begin{align}
f_\sigma = f_n \circ \sigma \circ f_{n-1} \circ \ldots \circ \sigma \circ f_1 \nonumber
\end{align}
\noindent where each $f_i : \mathbb{R}^{d_i} \rightarrow \mathbb{R}^{d_{i+1}}$ is an affine map (where $d_2,\ldots, d_{n-1}$ are the widths of the hidden layers of $f_\sigma$) and $\sigma (x_1,\ldots, x_l) \coloneqq (\max(0, x_1), \max(0, x_2),\ldots, \max(0, x_l))$ for any $1 \leq l \leq m$. $\blacksquare$

\subsection*{Framework}\label{sec:method}
\begin{figure}
\centering
\includegraphics[trim=0cm 0cm 0cm 0cm,clip=true,width=.9\textwidth]{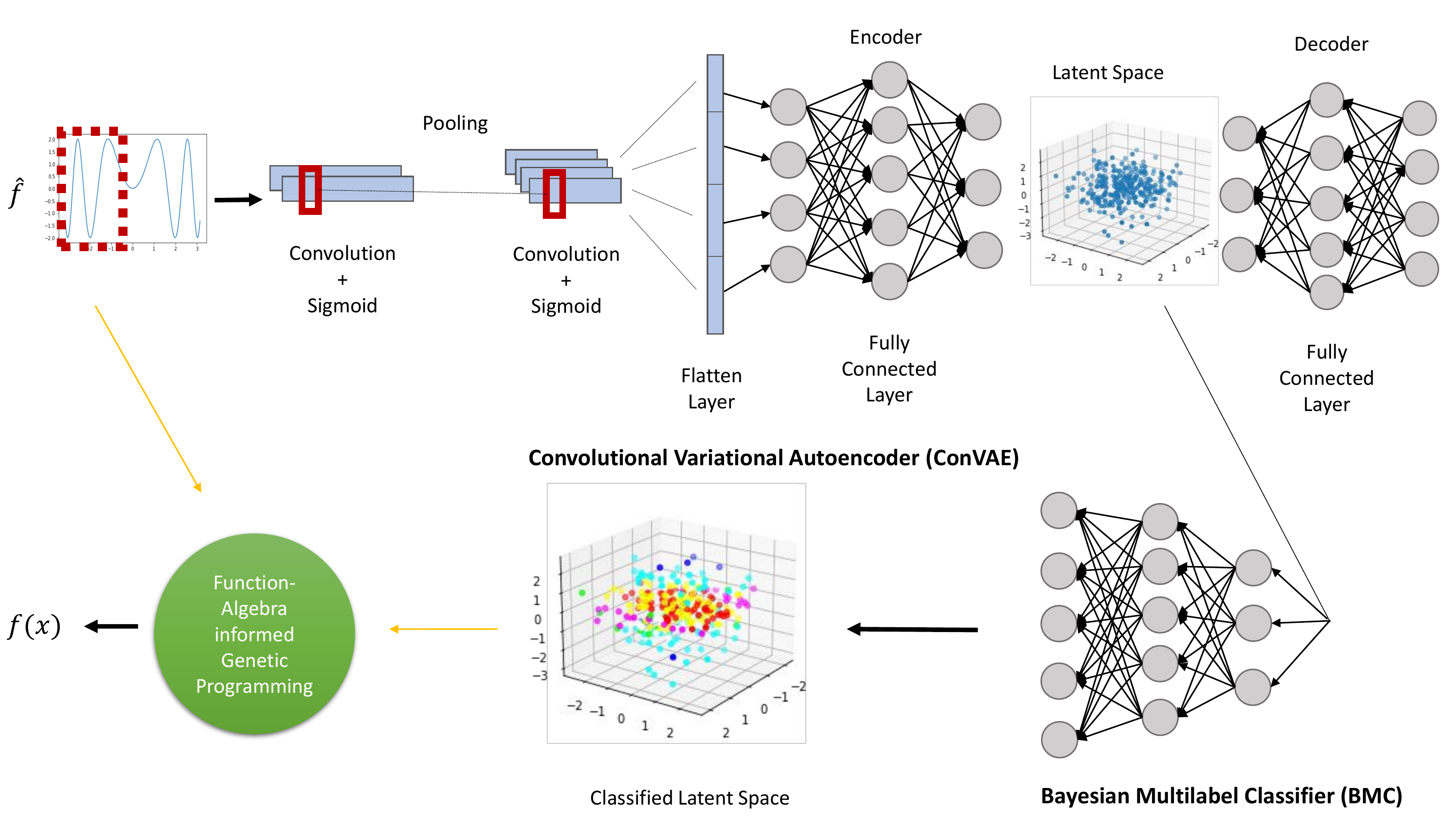}
\caption{\textbf{Overview of the framework.} A convolutional variational autoencoder (ConVAE) maps the input data onto a latent space. Here, the input (original) data is denoted as $\widehat{f}$. Then a Bayesian multilabel classifier (BMC), using the latent space from ConVAE, generates a posterior probability of an operator in the library ${\cal F}$ being used in the final tractable expression to recapitulate the observed data. Informed by the probability distribution from the BMC, the Function-Algebra-Informed Genetic Programmer evolves programs from a population of "unfit" programs (by means of an evolutionary process and through a new grammar) to identify the underlying mathematical expression. The prior distribution and the original data are the input (orange lines) to the genetic programmer.
}
\label{fig:workflow}
\end{figure}

Our overall framework consists of three sequential components (\textbf{Figure~\ref{fig:workflow}}). A ConVAE maps the input data onto a latent space. The probability distribution over the latent space is passed to a Bayesian multilabel classifier (BMC). The output of the classifier is a set of posterior probabilities, each of which indicates the probability that an element of the library is used in the final tractable expression to successfully reproduce the observed data. Finally, guided by the posterior probability from the BMC as the prior, faiGP aims to find a tractable expression that describes the data. It suffices for faiGP to restrict the search space to the set of functions that can be represented through the grammar $G$, as Theorem 1 guarantees that this algebra has the uniform approximation property.

It should be noted that the dimension of the ConVAE's latent space is related to the complexity of the underlying expression that describes the dataset. This complexity has to do with the number of distinct operators with which the final expression is constructed. \comment{At most, the number of unique operators is equivalent to the size of the library ${\cal F}$.}

\par\null\par
\noindent \textbf{Coefficients.} \label{subsec:coefficients} Fundamental constants in Nature (e.g., Newtonian constant of gravitation $G$, the mass of the electron, the fine-structure constant) appear in the physical sciences. They have achieved increasingly precise measurements over the years with technological and methodological advances. Notably, experimental deviation of a fundamental constant from theory is often pursued with interest because of its potential to break existing models.

In \cite{Koza1994}, Koza proposed an approach to discovering coefficients in symbolic regression applications to empirical data. The initial values are chosen from a predefined interval on the real axis. During the evolutionary process, these coefficients can be updated with new values, randomly chosen again, with a pre-defined probability. In this approach, the coefficients, variables, and functions in a symbolic expression are treated similarly. In contrast, since the values of the coefficients are drawn from an uncountably infinite set whereas variables and functions are chosen from a finite set, we handle the coefficients differently. In particular, we propose a hybrid coefficient-update system based on a random assignment and a curve-fitting algorithm. The former modifies the value of some of the coefficients, each randomly chosen with a pre-defined probability (here, 15\%), by a small amount randomly chosen from [-1, 1] during the point evolution; the latter, similarly to the former, first, randomly chooses some of the coefficients with a pre-defined probability (here, 15\%), and then fits a parametric curve or surface to the chosen coefficients during the evaluation of the expression to determine the best value for those coefficients. Here there is a trade-off between the time complexity of faiGP and the complexity of the final expression and its level of fitness, which, depending on the application, should be considered by the user. 

\par\null\par
\noindent \textbf{Loss function.} Evaluating how well faiGP models the given data is done through a loss function $\gamma: {\cal P} \rightarrow \mathbb{R}$, which quantifies the fitness of a proposed expression. Typically, there is no one-size-fits-all loss function. We test several loss functions, including $\chi^2$ loss, weighted Pearson, and mean absolute error, to determine the impact of the choice on performance. 

\par\null\par
\noindent \textbf{Regularizer.} \label{subsec:regularizer} To guide faiGP towards a desired symbolic expression of the underlying generative function, we can add regularization terms to the loss function. These regularization terms are usually application-dependent. A common regularization term used across all applications is a measure of expression complexity. This measure can be simply defined as the length of the expression. For example, between two expressions with identical level of fitness, the one with smaller length is considered the better candidate expression of the underlying generative function. 

Besides the complexity measure, we can add a second term that measures diversity among different generations. The guiding assumption is that diversity helps to ensure convergence of faiGP. Here we define two different measures of diversity, the first of which we adapt from analogous notions of specialization and diversity in the transcriptome \cite{Martnez9709}. We begin with the information-theoretic $H_j$, which is based on Shannon's entropy formula \cite{6773024}:

\begin{equation}
    H_j = - \sum_{i = 1}^{|\cal F|} p_{ij} \log_2(p_{ij})
    \label{eq:H_j}
\end{equation}

\noindent where $j$ is the index of an expression in a given generation and $|\cal F|$ is the cardinality of $\cal F$ (see production rules in (\ref{eq:grammar})), i.e., the number of available operators. $p_{ij}$ is the frequency of the $i$-th operator in the $j$-th expression. We can calculate the average frequency $p_i$ of the $i$-th operator among the $N$ expressions in a generation:

\begin{equation}
    p_i = \frac{1}{N} \sum_{j=1}^N p_{ij} \nonumber
    \label{eq:P_i}
\end{equation}

\noindent We define the following "specificity" measures:

\begin{equation}
    S_i = \frac{1}{N} \left( \sum_{j=1}^N \frac{p_{ij}}{p_i} \log_2 \frac{p_{ij}}{p_i} \right)
\end{equation}

\begin{equation}
    \delta_j = \sum_{i = 1}^{|\cal F|} p_{ij} S_i \nonumber
\end{equation}

\noindent Here $S_i$ provides a measure of the specificity of the $i$-th operator, quantifying the information on the identity of its source expression, and $\delta_j$ is the average operator specificity of the $j$-th expression. 

The first measure of diversity, $D_j$, is an analog of the KL divergence, quantifying the degree to which the $j$-th expression deviates from the distribution of the system:

\begin{equation}
    H_{Rj} = - \sum_{i = 1}^{|\cal F|} p_{ij} \log_2(p_i) \nonumber 
\end{equation}

\begin{equation}
    D_j = H_{Rj} - H_j.
\end{equation}

\noindent Although $D_j$ gives only an aggregate measure of diversity, its time complexity of $N*L$, where $L$ is the average length of an expression in a generation, makes it a valuable metric. A second measure of diversity can be defined based on the binary operator $\dot{\cap}$:

\begin{equation}
    U_1 ~\dot{\cap}~ U_2 = \begin{cases}
        |O_1 \cap O_2| ~~ \textrm{if} ~~ F_1 = F_2 \\
        0 ~~ \textrm{otherwise}
    \end{cases} \label{eq:cap}
\end{equation}

\begin{equation}
    \dot{D_j} = \sum_{k=1}^{N} E_j ~\dot{\cap}~ E_k
\end{equation}

\noindent where $E_j$ is the $j$-th expression in a generation. $\dot{D_j}$ gives a more granular measure of diversity in a generation at a higher time complexity of $N*N*L$. 

\section*{Results}
We evaluate the effect of different types of regularizers, including diversity (based on Shannon's entropy) and complexity (such as expression length). In addition, we test the performance of a purely stochastic as well as a hybrid (curve-fitting, model-based) approach to coefficient assignment. For all runs, we allow the expressions to evolve over 100 generations or until the loss function reaches a predefined value of 0.01. 

To investigate the complexity and accuracy of the resulting expression and the computational overhead, we perform a series of tests on some conventional benchmark functions (\textbf{Table~\ref{table:benchmarks}}) \cite{mundhenk2021symbolic}. We add ``Keijzer-2'' and ``Keijzer-2*'' \cite{keijzer2003improving}, the former corresponding to dropping ConVAE and BMC in generating the output, to the list to test performance in the presence of multiple local minima. Existing symbolic regression studies lack consistency in benchmarks and performance metric. In addition, most do not report the recovered symbolic expressions to facilitate reproducibility, making comparisons extremely challenging. Here, we provide ConVAE's reconstruction of the input, BMC's posterior inference, and faiGP's output to highlight some methodological points of general interest and the highly non-trivial relationships, which we term ``Ramanujan expressions,'' that can be generated from the combination of genetic programming and neural network implemented here.

\begin{table}
\begin{tabular}{lll}
\hline
\hline
Name & Expression & Dataset \\
\hline
Nguyen-1 & $x^3 + x^2 + x$ & $U(-1, 1, 20)$ \\
Nguyen-2 & $x^4 + x^3 + x^2 + x$ & $U(-1, 1, 20)$ \\
Nguyen-3 & $x^5 + x^4 + x^3 + x^2 + x$ & $U(-1, 1, 20)$ \\
Nguyen-4 & $x^6 + x^5 + x^4 + x^3 + x^2 + x$ & $U(-1, 1, 20)$ \\
Nguyen-5 & $\sin{(x^2)}\cos{(x)}-1$ & $U(-1, 1, 20)$ \\
Nguyen-6 & $\sin{(x)} + \sin{(x + x^2)}$ & $U(-1, 1, 20)$ \\
Nguyen-7 & $\log{(x + 1)} + \log{(x^2 + 1)}$ & $U(0, 2, 20)$ \\
Nguyen-8 & $\sqrt{x}$ & $U(0, 4, 20)$ \\
Nguyen-9 & $\sin{(x)} + \sin{(y^2)}$ & $U(0, 1, 20) \times U(0, 1, 20)$ \\
Nguyen-10 & $2 \sin{(x)} \cos{(y)}$ & $U(0, 1, 20) \times U(0, 1, 20)$ \\
Nguyen-11 & $x^y$ & $U(0, 1, 20) \times U(0, 1, 20)$ \\
Nguyen-12 & $x^4 - x^3 + \frac{1}{2}y^2 - y$ & $U(0, 1, 20) \times U(0, 1, 20)$ \\
Nguyen-$12^*$ & $x^4 - x^3 + \frac{1}{2}y^2 - y$ & $U(0, 10, 20) \times U(0, 10, 20)$ \\
\hline
R-1 & $\frac{(x+1)^3}{x^2-x+1}$ & $E(-1, 1, 20)$ \\
R-2 & $\frac{x^5-3x^3+1}{x^2+1}$ & $E(-1, 1, 20)$ \\
R-3 & $\frac{x^6+x^5}{x^4+x^3+x^2+x+1}$ & $E(-1, 1, 20)$ \\
R-$1^*$ & $\frac{(x+1)^3}{x^2-x+1}$ & $E(-10, 10, 20)$ \\
R-$2^*$ & $\frac{x^5-3x^3+1}{x^2+1}$ & $E(-10, 10, 20)$ \\
R-$3^*$ & $\frac{x^6+x^5}{x^4+x^3+x^2+x+1}$ & $E(-10, 10, 20)$ \\
\hline
Livermore-1 & $\frac{1}{3} + x + \sin{(x^2)}$ & $U(-10, 10, 1000)$ \\
Livermore-2 & $\sin{(x^2)}\cos{(x)}-2$ & $U(-1, 1, 20)$ \\
Livermore-3 & $\sin{(x^3)}\cos{(x^2)}-1$ & $U(-1, 1, 20)$ \\
Livermore-4 & $\log{(x+1)}+\log{(x^2+1)}+\log{(x)}$ & $U(0, 2, 20)$ \\
Livermore-5 & $x^4 - x^3 + x^2 - y$ & $U(0, 1, 20) \times U(0, 1, 20)$ \\
Livermore-6 & $4x^4 + 3x^3 + 2x^2 + x$ & $U(-1, 1, 20)$ \\
Livermore-7 & $\sinh{(x)}$ & $U(-1, 1, 20)$ \\
Livermore-8 & $\cosh{(x)}$ & $U(-1, 1, 20)$ \\
Livermore-9 & $x^9 + x^8 + x^7 + x^6 + x^5 + x^4 + x^3 + x^2 + x$ & $U(-1, 1, 20)$ \\
Livermore-10 & $6\sin{(x)}\cos{(y)}$ & $U(0, 1, 20) \times U(0, 1, 20)$ \\
Livermore-11 & $\frac{x^2 y^2}{x+y}$ & $U(-1, 1, 50) \times U(-1, 1, 50)$ \\
Livermore-12 & $\frac{x^5}{y^3}$ & $U(-1, 1, 50) \times U(-1, 1, 50)$ \\
Livermore-13 & $x^\frac{1}{3}$ & $U(0, 4, 20)$ \\
Livermore-14 & $x^3 + x^2 + x + \sin{(x)} + \sin{(x^2)}$ & $U(-1, 1, 20)$ \\
Livermore-15 & $x^\frac{1}{5}$ & $U(0, 4, 20)$ \\
Livermore-16 & $x^\frac{2}{5}$ & $U(0, 4, 20)$ \\
Livermore-17 & $4\sin{(x)}\cos{(y)}$ & $U(0, 1, 20) \times U(0, 1, 20)$ \\
Livermore-18 & $\sin{(x^2)}\cos{(x)}-5$ & $U(-1, 1, 20)$ \\
Livermore-19 & $x^5 + x^4 + x^2 + x$ & $U(-1, 1, 20)$ \\
Livermore-20 & $\exp{(-x^2)}$ & $U(-1, 1, 20)$ \\
Livermore-21 & $x^8 + x^7 + x^6 + x^5 + x^4 + x^3 + x^2 + x$ & $U(-1, 1, 20)$ \\
Livermore-22 & $\exp{(- 0.5x^2)}$ & $U(-1, 1, 20)$ \\
\hline
Keijzer-2 & $0.3 x \sin{(2 \pi x)}$ & $U(-2, 2, 300)$ \\
Keijzer-2* & $0.3 x \sin{(2 \pi x)}$ & $U(-2, 2, 20)$ \\
\hline
\hline
\end{tabular}\caption{\textbf{The set of benchmark functions and the corresponding datasets}. $U(a, b, n)$ refers to $n$ random draws from the continuous uniform distribution $U(a, b)$ defined by the bounds $a$ and $b$, where $b > a$. $E(a, b, n)$ denotes the set of fixed boundary points defined by $n$ equally-spaced segments within the interval $(a,b)$. A benchmark function $f(X)$ is either a function of a single random variable $X$ where $X$ follows the probability distribution $U(a, b)$ from which $n$ points are drawn to generate the input dataset or a function of a variable defined at the fixed set $E(a, b, n)$. In some cases, a benchmark function is a function $f(X, Y)$  of two random variables $X$ and $Y$, with functional values generated from the product $U(a, b, n) \times U(a, b, n)$. ``Keijzer-2'' and ``Keijzer-2*'', which test performance in the presence of multiple local minima, are from reference \cite{keijzer2003improving} but with a modified dataset.}
\label{table:benchmarks}
\end{table}

\begin{table}
    \centering
    \begin{tabular}{llll}
        Loss Function & Average Expression Length & Average Run Time & $R^2$ \\
        \hline
        Mean Absolute Error & $3.13 \pm 0.43$ & $192.94 \pm 28.55$ & $-0.01 \pm 0.12$ \\
        Mean Square Error & $3.2 \pm 2.21$ & $151.89 \pm 60.02$ & $0.25 \pm 0.19$ \\
        Root Mean Square Error & $5.57 \pm 5.4$ & $138.46 \pm 28.11$ & $0.11 \pm 0.18$ \\
        Pearson & $12.43 \pm 3.50$ & $8.29 \pm 0.97$ & $0.31 \pm 0.15$ \\
        Spearman & $12.80 \pm 3.08$ & $9.32 \pm 1.72$ & $0.21 \pm 0.32$ \\
        $\chi^2$ & $41.53\pm14.98$ & $4091.72 \pm 2358.7$ & $0.95 \pm 0.1$ \\
    \end{tabular}
    \caption{\textbf{Loss function evaluation.} A performance comparison of different loss functions used in faiGP on ``Keijzer-2'' benchmark. The reported results are the average over 30 runs. Expression length is calculated according to equation (\ref{eq:length_with_p}). Run time is measured in seconds.}
    \label{table:loss_function}
\end{table}

\begin{table}
\scriptsize
\begin{tabular}{p{0.11\linewidth} | p{0.78\linewidth} | p{0.05\linewidth} | p{0.06\linewidth}}
\hline
\hline
Name & Recovered Expression  & $R^2$ & Search Time (sec) \\
\hline
Nguyen-1 & $x^3 + x^2 + x$ & 1.00 & 4.37 \\
Nguyen-2 & $x^4 + x^3 + x^2 + x$ & 1.00 & 3.98 \\
Nguyen-3 & $0.76(x + (0.23x + 0.72*0.72x + (0.7x)^3 + 0.92\sqrt{0.56x})^3)$ & 1.00 & 7.24 \\
Nguyen-4 & $x + 0.59x^2 + 0.78x^4 + 0.15 (0.84(0.85\sin{(x)} + x^2 + \sqrt{x}))^4 + 0.16 \sin{(x)}$ & 1.00 & 10.51\\
Nguyen-5 & $0.44(-0.74 + 0.26\log{(0.06x)} + (0.83(\log{(0.48x)} \sqrt{x}))^5 + 0.53\sqrt{0.53x})$ & 1.00 & 8.68 \\
Nguyen-6 & $\sin{(-0.52(0.75x)^5 + 0.99x + 0.96(0.99x)^2)} + \sin{(x)}$ & 1.00 & 10.94 \\
Nguyen-7 & $0.58x^2 + x - 0.21x^3$ & 1.00 & 4.67 \\
Nguyen-8 & $\sqrt{x}$ & 1.00 & 4.53 \\
Nguyen-9 & $0.88\cos{(0.24x)} + \log{(y + 0.49(\cos{0.75y + (\cos{(0.75y)} + 0.92y + \sin{(y)})} + \sqrt{y}))} + \sin{(x)}$ & 1.00 & 7.78 \\
Nguyen-10 & $-0.48(\log^3{(-0.17y)} \sin{(x)} 0.52\sin{(y)})$ & 1.00 & 85.69 \\
Nguyen-11 & $0.78(-0.38y - 0.28\log{(0.57y)} + 0.88\sqrt{0.72x} + 0.74 x y)$ & 1.00 & 176.41 \\
Nguyen-12 & $-0.069(4.110y-0.176x+3.625\sin{(2.108y)}-2.730(x^2 \cos{(1.532x)} \log{(0.075y)} \sqrt{1.12x}))$ & 0.98 & 24.11 \\
Nguyen-$12^*$ & $0.66 x^3 \sqrt{x} + 0.02 x^3 \sin{(x)} + 0.22 x^4 \sqrt{x}$ & 1.00 & 24.63 \\
\hline
R-1 & $0.939(x+0.84+(0.831(x)+\sqrt{x})^2+\sin{(x)}+0.965(0.603x-0.889(0.868x)^2+\sin{(0.982x)}+\sqrt{0.997x})^2)$ & 1.00 & 10.73 \\
R-2 & $\sin{(-0.484x)}+\sin{(-0.376\log{(x)}+0.989\sin{(-x)}+0.222+\sin{(-0.799\log{(0.513x)})})}$ & 1.00 & 57.08 \\
R-3 & $0.242(-0.455(0.413x)^2+(0.996x)^6+(0.998x)^3)$ & 1.00 & 15.68 \\
R-1* & $3.947+x+\sin{(\sin{(\frac{0.994}{0.41x})})}+\frac{0.661}{0.452x}+\sin{(\frac{0.985}{0.41x})}$ & 1.00 & 514.38 \\
R-2* & $0.954(-0.24-0.656(x)+(x)^3-0.752\sqrt{x})$ & 1.00 & 1.72 \\
R-3* & $-0.71+(0.997x)^2+0.16\sin{(x)}$ & 1.00 & 1.72 \\
\hline
Livermore-1 & $x+0.334+0.983\sin{(x^2)}$ & 1.00 & 2931.70 \\
Livermore-2 & $0.603((0.074x)^2+(0.438(0.498\log{((0.631x)^2)}-\sqrt{0.108x}))^2+\log{(0.075x)})$ & 1.00 & 10.31 \\
Livermore-3 & $-0.98-0.02\cos{(-0.8x)}-0.648(0.916x -(\sqrt{0.919x})^2)-0.581(0.919x -0.145\sqrt{-0.053x})$ & 1.00 & 40.18 \\
Livermore-4 & $0.96(x-0.09(-0.787x)^4+(0.487x)^2+\log{(0.79x)}+0.473\sqrt{x})$ & 1.00 & 8.06 \\
Livermore-5 & $0.173x - y - 0.011 x y + 0.819x^3$ & 1.00 & 168.78 \\
Livermore-6 & $0.999((x+x^3+0.588(0.617x)^2+0.826\sqrt{x})^2+x^4)$ & 1.00 & 4.17 \\
Livermore-7 & $0.334x^3+\sin{(x)}$ & 1.00 & 4.11 \\
Livermore-8 & $0.892(0.85x)^2+\cos{(0.478x)}$ & 1.00 & 3.51 \\
Livermore-9 & $x+(-0.937x)^6+0.205\sqrt{x}+0.324x (0.965(0.843x+0.856x+(0.254x)^3+0.472\sqrt{x}))^4$ & 1.00 & 3.75 \\
Livermore-10 & $0.726(0.99\log^2{(0.14y)}\sin{(0.837x)} \sqrt{0.544y}+0.982\sin{(x)}+0.711\cos{(y)} \log^2{(0.139y)} \sin{(0.999x)}\sqrt{y} - 0.185\cos{(0.645y)}\sin{(0.513x)})$ & 1.00 & 19.75 \\
Livermore-11 & $0.145(y \log{(0.258y+0.468\log{(0.976x)}-0.488\log{(0.37\log{(0.458x)})})} \log{(0.446x+0.338y)})$ & -0.04 & 634.49 \\
Livermore-12 & $\log{(-0.659(\cos{(-0.342x)}+\sin{(y)}+\log{(y)} \sqrt{0.379(y+y+\cos{(x)}+0.005(x)+\sqrt{x})}))}+\sin{(y)}$ & -0.06 & 4.35 \\
Livermore-13 & $\sqrt{0.313(x)+\sqrt{0.438x}}$ & 1.00 & 6.60 \\
Livermore-14 & $0.999(2.12x+0.171\sqrt{-0.004x}+0.973x^2+0.344x\sin{(x)}+0.997x^2\sin{(0.371x)}+0.819x\sin{(0.754x)}-0.012\cos{(0.962x)} \sin{(x)}+0.499\sin{(0.597x)} \cos{(\sin{(-0.138x)})}+0.574\log{(x)} \sin{(0.445x)})$ & 1.00 & 22.60 \\
Livermore-15 & $0.17+0.138x-0.149x+\sqrt{0.712\sqrt{x}}$ & 1.00 & 5.60 \\
Livermore-16 & $-1.09x+0.917\sqrt{-4.114(0.003(9.935x)^2+0.996\sqrt{0.461(x^2+1.123x^3)})}
$ & 1.00 & 12.56 \\
Livermore-17 & $0.981x+0.682x+0.861(0.603(x) 0.861\log{(0.023x)} \sin{(\log{(0.786y)})})$ & 1.00 & 11.55 \\
Livermore-18 & $0.895(0.032(-0.779x)+(x)^2+(0.037x)^6+0.84 \log{(0.117x)}-0.317(-0.484\log{(x)} \sin{(0.014(0.0x)^4+\log{(-0.026x)})})+\log{(-0.037x)} 0.99\sqrt{x})$ & 0.96 & 11.55 \\
Livermore-19 & $(0.759x+0.502(x)^2)^3+0.589x+x^2+0.373\sin{(x)}$ & 1.00 & 7.00 \\
Livermore-20 & $-0.668(0.97\log{(x)}+0.152(0.97\log{(0.997x)}-0.302x^2+0.85\sin{(0.971\log{(0.128x)}+0.757\sqrt{x})})^2+0.85\sin{(0.88\log{(0.128x)}+0.787\sqrt{x})})$ & 1.00 & 9.53 \\
Livermore-21 & $0.977(0.811(x)-0.427\log{-0.045x\log{(0.044x)}))}(x+0.997(0.986x)^2+x^5)^2)$ & 1.00 & 5.59 \\
Livermore-22 & $(0.514x)^4+\cos{(x)}$ & 1.00 & 6.37 \\
\hline
Keijzer-2  & $0.302 (x  \sin{(0.979 \log^2{(0.257 x)} x \sqrt{x})}  \cos{0.996(0.990(0.931x)^2 - 0.153\log^2{(0.005x)} \cos{(0.982x)})})$ & 1.00 & 1209.56 \\
Keijzer-2* & $-0.300 x \sin{(-6.283x)}$ & 1.00 & 11.66 \\
\hline
\hline
\end{tabular}\caption{\textbf{Benchmark results.} For each benchmark function, the recovered expression, the $R^2$, and the search time (in seconds) are presented. We report the maximum $R^2$ from 10 solutions to each benchmark function. Notably, the recovered expressions can present some very interesting (i.e., non-trivial) equalities or approximations. For example, $\exp{(- 0.5x^2)} \approx (0.514x)^4+\cos{(x)}$  (see ``Livermore-22'') and $\sinh{(x)} \approx 0.334(x)^3+\sin{(x)}$ (see ``Livermore-7''), each side defined on $U(-1, 1, 20)$ and each showing Spearman correlation $\rho = 1$ (as can also be seen from the near-perfect diagonal scatterplot) between the two sets of functional values (i.e., original and recovered). By definition, the operator $\sqrt{}$ refers to the square root of the absolute value and the $\log$ operator refers to the natural log of the absolute value, with $\log(0) \coloneqq 0$.}
\label{table:benchmarks_results}
\end{table}

\par\null\par
\begin{figure}
\centering
\includegraphics[trim=0cm 1cm 0cm 0cm,clip=true,width=.8\textwidth,height=.35\textwidth]{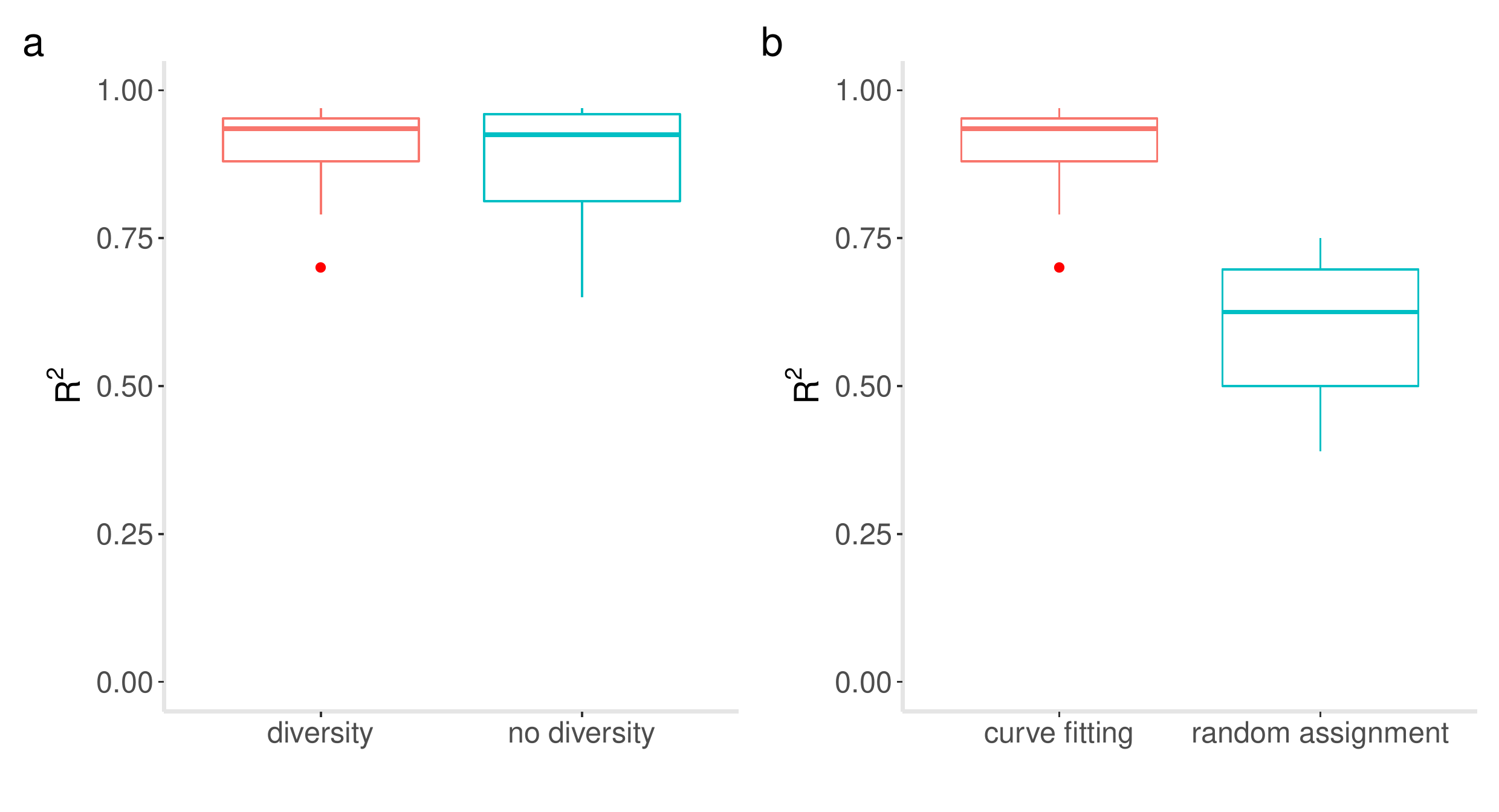}
\caption{\textbf{Implementation choices.} Using the ``Keijzer-2'' benchmark, we investigate the impact of design choices on the fitness score. \textbf{a)} Utilizing a diversity regularizer can lead to substantially reduced variance in the fitness score $R^2$, even when there is no significant difference in the mean fitness score, relative to absence of a diversity regularizer. \textbf{b)} A significant gain in $R^2$ can be obtained by leveraging a curve-fitting procedure to determine the coefficients relative to random assignment.}
\label{fig:comparisons_regularizer}
\end{figure}

\noindent \textbf{Network training.} We train our networks, ConVAE and BMC, with a dataset generated from a pre-selected library ${\cal F}$. The library is not unique and, depending on what operators we allow to be part of the final expression (in Rule 7 of the grammar), can include more or fewer operators. Starting with ConVAE, over 20 epochs, we map the generated dataset onto a latent space. As we previously noted, the size of the latent space is related to the complexity of the expression and can change based on user requirements. 

The training dataset for ConVAE and BMC consists of 3500 randomly generated expressions for each $f$ in the library ${\cal F} = \{\sqrt{}, \cos, \sin, \log, \mathrm{af\!fine}\}$. Each expression $y=y(x)$ is generated by randomly choosing 4 constants $a$, $b$, $c$, and $d$ in $\mathbb{R}$:

\begin{equation}
	y(x) = a*f(b*x+c)+d. \label{eq:NT1}
\end{equation} 

\noindent for each $f \in {\cal F}$ and $x$ in a fixed closed interval $[x_{\textrm{min}}, x_{\textrm{max}}]$.

We train the BMC network on ConVAE's latent space with the known labels that associate values in the latent space with the specific operators used to generate the training dataset. For BMC, the parameters $\phi$ of the parametric distribution $q_\phi(\bm{w})$ (equation (\ref{eq:BMC_likelihood})) are learned by a forward pass neural network. The log likelihood ${\cal L}_{\cal D}(\phi)$ is estimated via Monte Carlo \cite{kingma2015variational}, and KL divergence is added as a regularization term \cite{molchanov2017variational}. BMC outputs a probability distribution over ${\cal F}$:

\begin{equation}
p(O | \bm{z}, {\cal D}) = \int p(O | \bm{z}, \bm{w}) p(\bm{w} | {\cal D})d\bm{w}
\label{eq:posterior}
\end{equation}

\noindent where $O \in \cal F$ is a label (operator), $\bm{z}$ is the ConVAE latent space, and $\cal D$ denotes the data. The posterior probability $p(O | \bm{z}, {\cal D})$ is generated by marginalizing over the weights $\bm{w}$ of the neural network. This posterior probability is then used to inform faiGP during the expression generation and discovery. 


%
\par\null\par

\begin{figure}
\centering
\includegraphics[width=1\textwidth, height=0.5\textwidth]{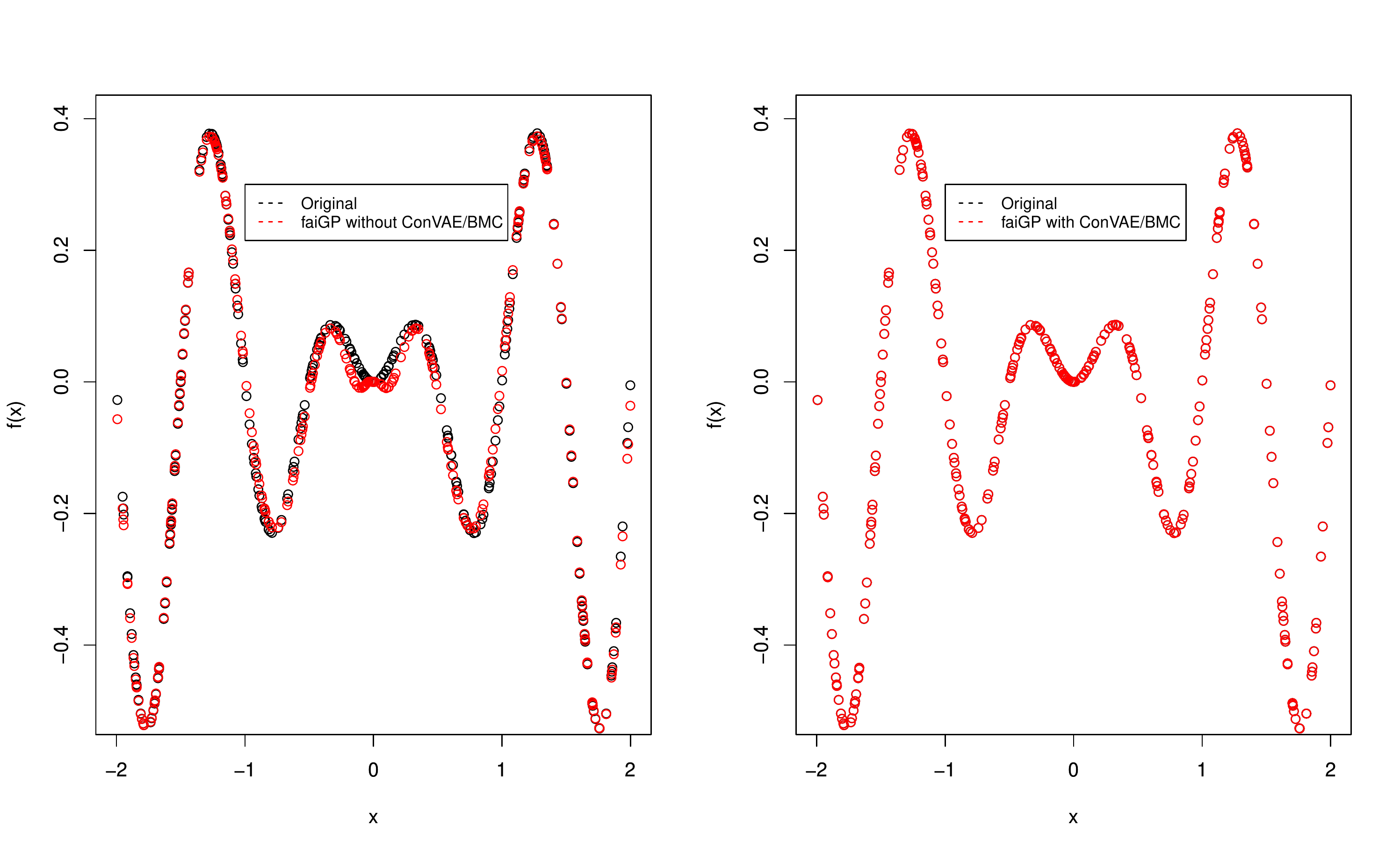}
\caption{\textbf{Leveraging ConVAE and BMC improves concordance.} Comparison of the original input (black), defined by the function $f(x) = 0.3 x \sin{(2 \pi x)}$, and faiGP output (red) on $U(-2, 2, 300)$. Note the irrational number $\pi$ in the operand for the sine function, which would require an approximating function in the output. Note also the presence of multiple local minima. Using ConVAE (for dimensionality reduction to reduce the search space) and BMC (for posterior inference to generate the prior for the genetic programmer) to direct the search leads to noticeable gain in performance in comparison with just using the genetic programmer. The right panel (``Keijzer-2*'' benchmark with output $-0.300 x \sin{(-6.283x)}$) shows better concordance than the left panel (``Keijzer-2'' benchmark with output $0.302 (x  \sin{(0.979 \log^2{(0.257 x)} x \sqrt{x})}  \cos{0.996(0.990(0.931x)^2 - 0.153\log^2{(0.005x)} \cos{(0.982x)})})$). Note the output $-0.300 x \sin{(-6.283x)}$ on the right is an even function, i.e., $f(x) = f(-x)$, hence, $-0.300 x \sin{(-6.283x)}$ = $0.300 x \sin{(6.283x)}$.}
\refstepcounter{SIfig}
\label{fig:VUMC_benchmark}
\end{figure}

\noindent \textbf{Loss function.} We find that the choice of the loss function can have a significant impact (\textbf{Table \ref{table:loss_function}}), both in the level of fitness and the complexity of the resulting symbolic expression. For example, on average, evaluated on the ``Keijzer-2" benchmark (\textbf{Figure~\ref{fig:VUMC_benchmark}}), $\chi^2$ results in the highest fitness ($R^2$) but also generates the greatest level of expression complexity (length) and requires the largest computational cost (run time). On the other hand, mean absolute error leads to poor fitness ($R^2 < 0$).

\par\null\par

\begin{figure}
    \centering
    \includegraphics[angle=-90,width=1\textwidth]{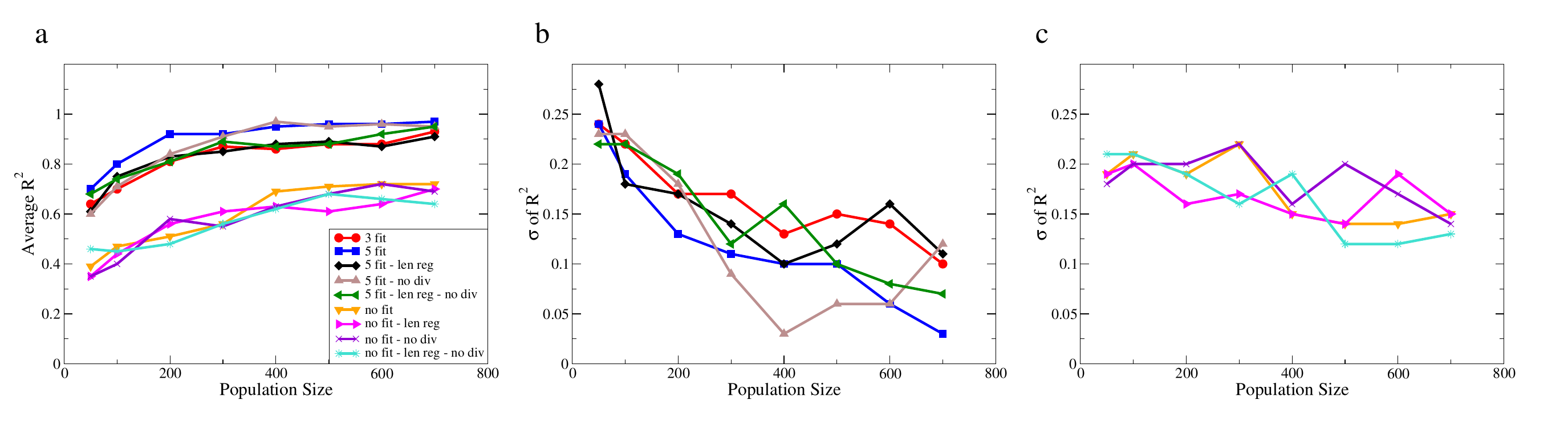}
    \caption{\textbf{Fitness.} Averaged over 30 runs on the ``Keijzer-2'' benchmark, \textbf{a)} $R^2$ as a function of population size. Standard deviation ($\sigma$) of $R^2$ \textbf{b)} with and \textbf{c)} without curve fitting, as a function of population size. ``3 fit'' indicates the curve fitting procedure with a maximum limit of 3 call backs for evaluating the expression. ``5 fit'' is similar to ``3 fit'' with a maximum of 5 calls. ``No fit'' indicates a run where curve fitting is not employed. ``Len reg'' indicates inclusion of a second length regularizer (equation (\ref{eq:length_no_p})), which penalizes longer expressions based on the length of expression as represented by the grammar \ref{eq:grammar}. Finally, ``no div'' indicates exclusion of Shannon's diversity regularizer.}
    \label{fig:comparisons_performance_R2}
\end{figure}

\begin{figure}
\centering
\includegraphics[angle=-90,width=1\textwidth]{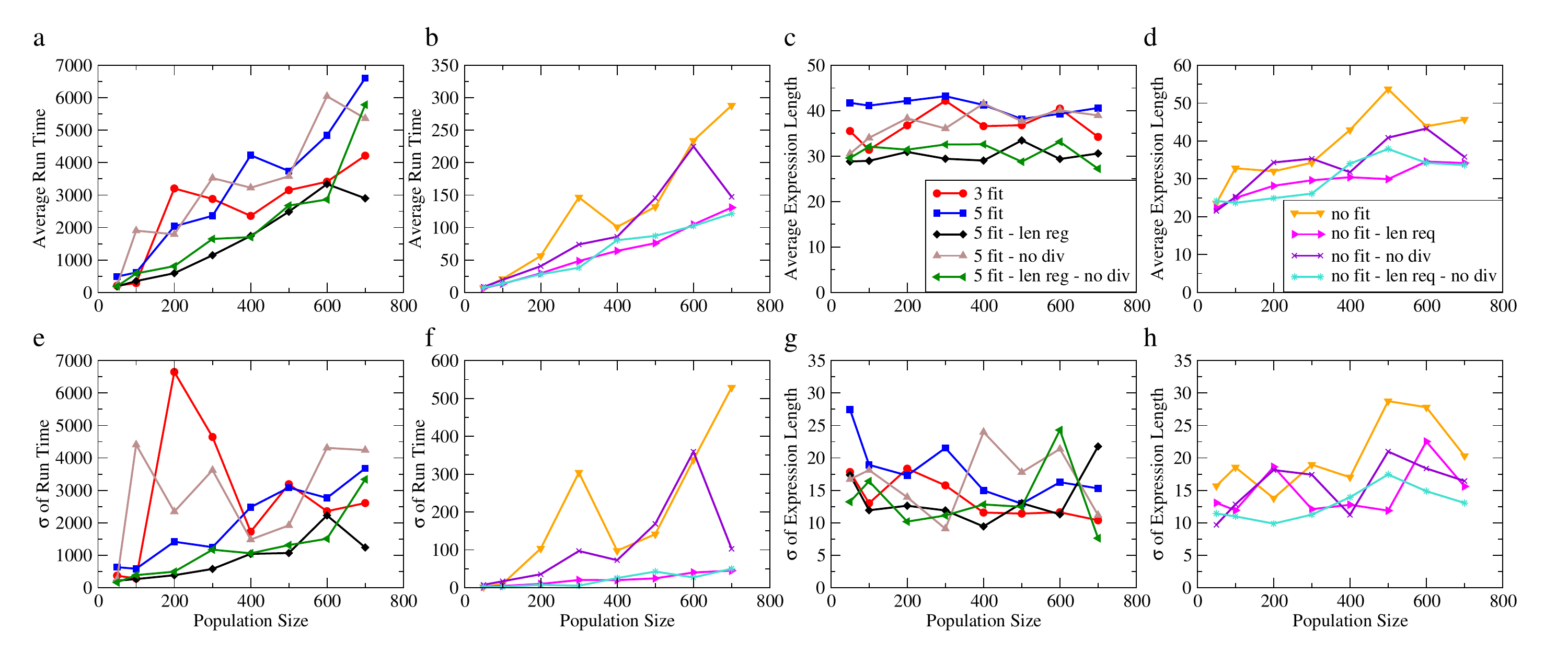}
\caption{\textbf{Run time and expression length.} Averaged over 30 runs on the ``Keijzer-2'' benchmark, \textbf{a)} run time in seconds with curve fitting \textbf{b)} run time in seconds without curve fitting \textbf{c)} average expression length with curve fitting \textbf{d)} average expression length without curve fitting, as functions of population size. \textbf{e, f)} Standard deviation of run time and \textbf{g, h)} standard deviation of expression length, as functions of population size. Expression length is calculated according to equation (\ref{eq:length_with_p}).}
\label{fig:comparisons_performance}
\end{figure}

\noindent \textbf{Regularization.} We find that leveraging a diversity regularizer can lead to significantly lower variability in the fitness score $R^2$, even when there is no significant difference in the mean of the fitness score, in comparison with leaving out the diversity regularizer (\textbf{Figure~\ref{fig:comparisons_regularizer}a}). Considering that diversity for a symbolic expression is not standardized, how diversity is \textit{estimated} may have a substantial impact on the final results.

We examine how the choice of estimator $\widehat{p_{ij}}$ of the true frequency $p_{ij}$ in the Shannon's entropy regularizer (equation (\ref{eq:H_j})) determines the resulting expression. One natural estimator is the maximum likelihood estimator, that is, the proportion of times the $i$-th operator is actually observed in the $j$-th expression. This estimator, however, results in expressions with a large number of zeros as exponents. An expression such as $[f(x)]^0$, where $f \in {\cal F}$, is mathematically equivalent to the constant expression 1 but is assigned a larger diversity measure. Thus, within the regularizer, this estimator, while asymptotically efficient \cite{van2000asymptotic}, would promote generation of bloat. Below, we conduct empirical studies of combinations of the diversity and complexity regularizers. 

\par\null\par

\noindent \textbf{Determining the coefficients.} We investigate the effect of using a curve-fitting algorithm within a probabilistic setup for coefficient assignment. We use the Levenberg–Marquardt algorithm for non-linear least squares curve fitting. Here we assume a limit (set to 3 and 5) on the maximum number of calls to the function during each fit. We find that a significantly greater fitness score $R^2$ can be obtained by utilizing the curve-fitting procedure than by a random (purely probabilistic) assignment (\textbf{Figure~\ref{fig:comparisons_regularizer}b}). Thus, probabilistically integrating an explicit model for fitting coefficients into the evolutionary process can lead to substantial performance gain. This gain in fitness score is passed on to the next generation.

\par\null\par

\begin{figure}
\centering
\includegraphics[angle=-90,width=1\textwidth]{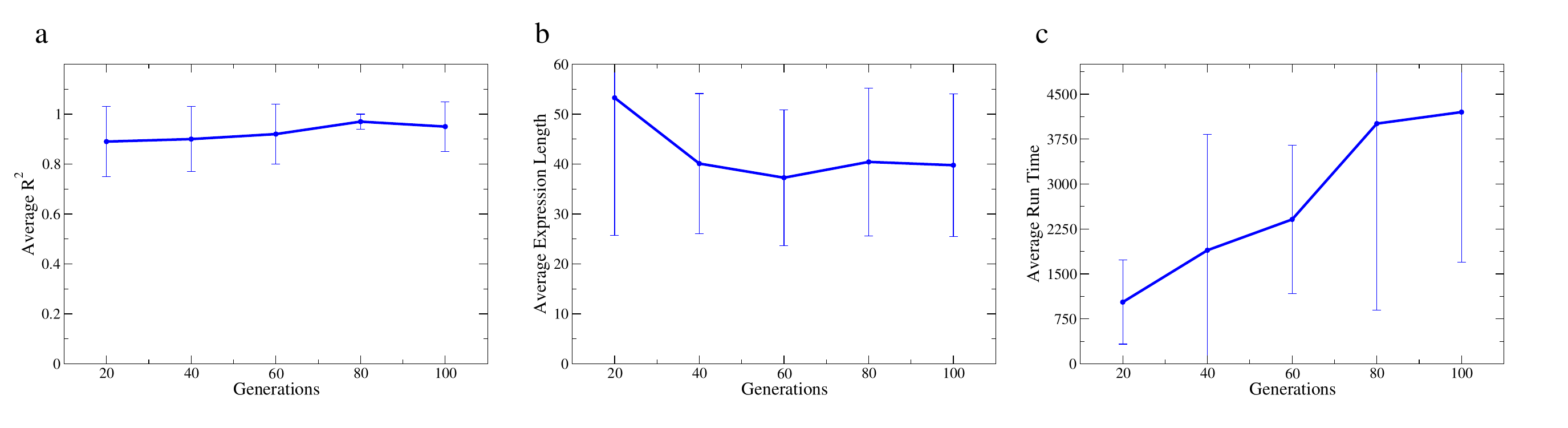}
\caption{\textbf{Number of generations.} Dependence of \textbf{a)} $R^2$ \textbf{b)} expression length and \textbf{c)} run time on number of generations using the ``Keijzer-2'' benchmark.}
\refstepcounter{SIfig}
\label{fig:gens}
\end{figure}

\noindent \textbf{Computational performance and fitness.} We collect the average value of $R^2$, the final expression length, and total run time over 30 runs as a function of population size (\textbf{Figure~\ref{fig:comparisons_performance_R2}} and \textbf{Figure~\ref{fig:comparisons_performance}}). Our finding (\textbf{Figure~\ref{fig:comparisons_performance_R2}a}) suggests faiGP can achieve an optimal solution with a population size of 400 in each generation. We thus also consider how many generations, on average, are required for faiGP to find an optimal solution while the population size is fixed at 400 in each generation. We examine the dependence of the metrics on the number of generations. Although expression length decreases (and run time substantially increases) with number of generations, the fitness score $R^2$ appears to be less dependent (\textbf{Figure~\ref{fig:gens}}).

We find that for each combination of regularizers, the probabilistic curve-fitting procedure outperforms random assignment. In addition, the significant gain in fitness is maintained at each population size tested (\textbf{Figure~\ref{fig:comparisons_performance_R2}}). We also evaluate the computational performance from the use of the diversity and complexity regularizers. For the complexity regularizer, we use two definitions of length:

\begin{equation}
    \mathrm{len}(\{(C, F, O, P)\}) = 3 + |O|  \label{eq:length_no_p}
\end{equation}
\noindent and 
\begin{equation}    
    \mathrm{len}(\{(C, F, {\cal E}, P)\}) = 1 + (1 + \mathrm{len}({\cal E})) \times P \label{eq:length_with_p}
\end{equation}

\begin{figure}
\centering
\includegraphics[angle=-90,width=1\textwidth]{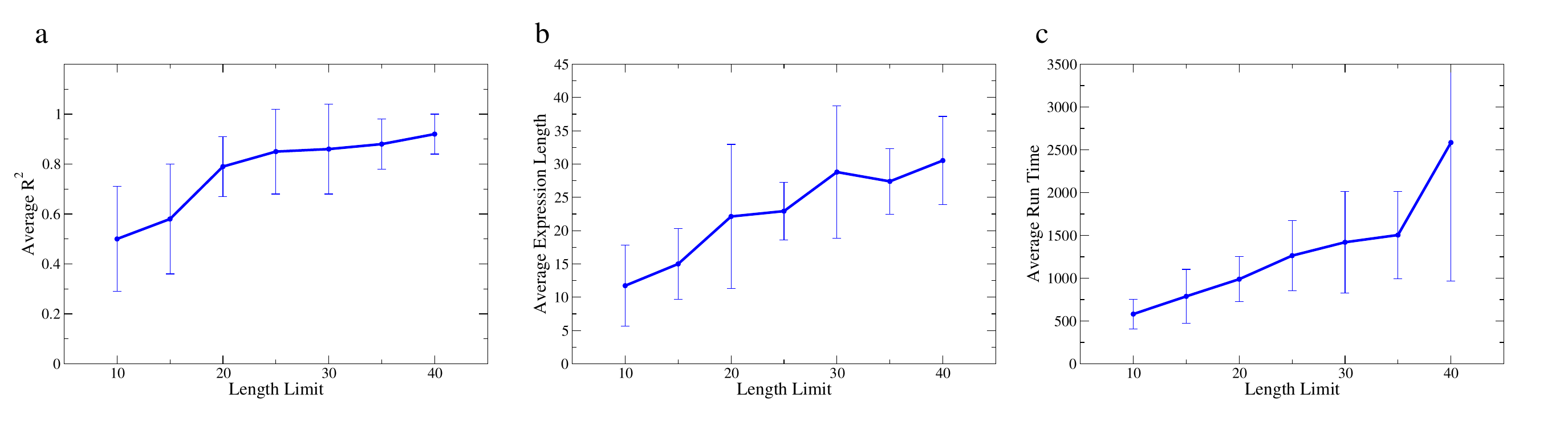}
\caption{\textbf{Expression length limit.} Impact of inclusion of an additional threshold regularizer $M * \mathds{1}_{\mathrm{len}(\{(C, F, O, P)\}) > \lambda}$, where $M \gg 0$ is the penalty and $\lambda$ is the length threshold, on \textbf{a)} $R^2$ \textbf{b)} expression length and \textbf{c)} run time using the ``Keijzer-2'' benchmark.}
\refstepcounter{SIfig}
\label{fig:length_limit}
\end{figure}

\noindent Here, $|O|$ is the cardinality of $O$, $\cal E$ is either $O$ or $S$ (see rule 6 in equation (\ref{eq:grammar})), and $\mathrm{len}(\cal E)$ is defined recursively. As regularizers, both definitions favor a less complex operator. The second definition, in addition, discourages learning a high value for the exponent $P$. In our analysis, this definition is used universally for regularization. We find that the computational overhead from ensuring diversity is minimal but is substantial from having to control the complexity of the expression. When we enforce a hard penalty $M \gg 0$ for expressions that exceed a pre-defined threshold $\lambda$, via inclusion of an additional threshold regularizer $M * \mathds{1}_{\mathrm{len}(\{(C, F, O, P)\}) > \lambda}$, we observe higher fitness score $R^2$, final expression length, and run time with higher threshold (see \textbf{Figure~\ref{fig:length_limit}} for the point estimate and variability at each value of $\lambda$). 

\begin{figure}
    \centering
    \includegraphics[trim=0cm 7cm 0cm 6.5cm,clip=true,width=.9\textwidth]{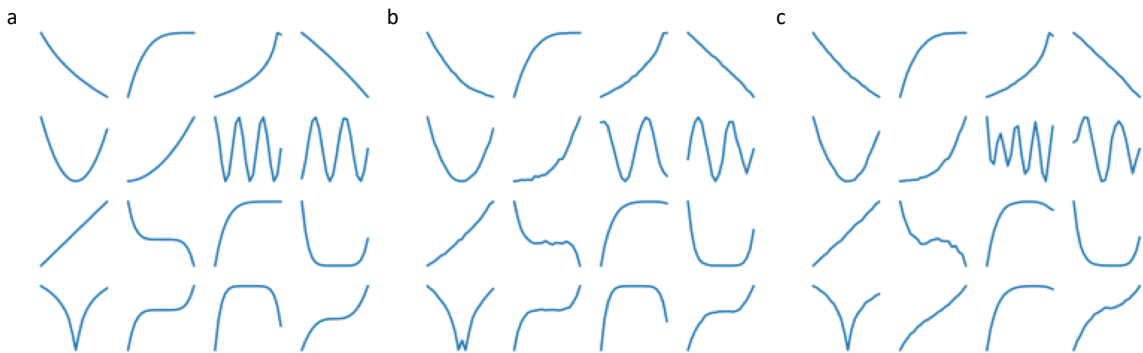}
    \caption{\textbf{ConVAE performance.} Over 1500 epochs, ConVAE is trained with data generated using equation (\ref{eq:NT1}) and the library ${\cal F}$. \textbf{a)} a set of sampled input data fed into the ConVAE encoder, each representing a set of ordered pairs $\{(x, y(x)) \} \subset \mathbb{R}^2$ for some $y(x)$ (equation (\ref{eq:NT1})) and \textbf{b, c)} two sampled outputs from the ConVAE decoder.}
    \label{fig:sampled_functions}
\end{figure}

\par\null\par

\noindent \textbf{BMC posterior inference using ConVAE latent space.} We find that ConVAE performs well in identifying a lower-dimensional latent space (see \textbf{Figure~\ref{fig:sampled_functions}} for comparisons of sampled input data and reconstructed output). The BMC posterior informs the initial search by capturing a key contributing operator of the underlying generative function (\textbf{Figure~\ref{fig:bmc_output}}). Leveraging ConVAE and BMC to reduce the search space and generate the prior for the genetic programmer (from the posterior inference using the ConVAE latent space), respectively, improves performance (\textbf{Figure~\ref{fig:VUMC_benchmark}}) in comparison with the baseline approach of passing a uniform prior to the genetic programmer.

\par\null\par

\noindent \textbf{Benchmarks.} We conduct comprehensive evaluation of the performance of faiGP, using a broad set of benchmark functions (\textbf{Table~\ref{table:benchmarks}}). Here, benchmarks are generative functions of one or two variables, serving as ground-truth expressions. The input of a benchmark function $f(X)$ of a single variable $X$ is generated from $n$ draws from the uniform distribution $U(a, b)$ (in this case, the input is denoted by $U(a, b, n)$) or is derived from a fixed set of points defined by $n$ equally-spaced segments within the interval $(a,b)$ (here, the input is denoted by $E(a, b, n)$). A benchmark function $f(X, Y)$  of two random variables $X$ and $Y$ generates a dataset of functional values from the product $U(a, b, n) \times U(a, b, n)$.

 \par\null\par
\begin{figure}
\centering
\includegraphics[trim=0cm 0cm 0cm 0cm,clip=true,width=0.9\textwidth]{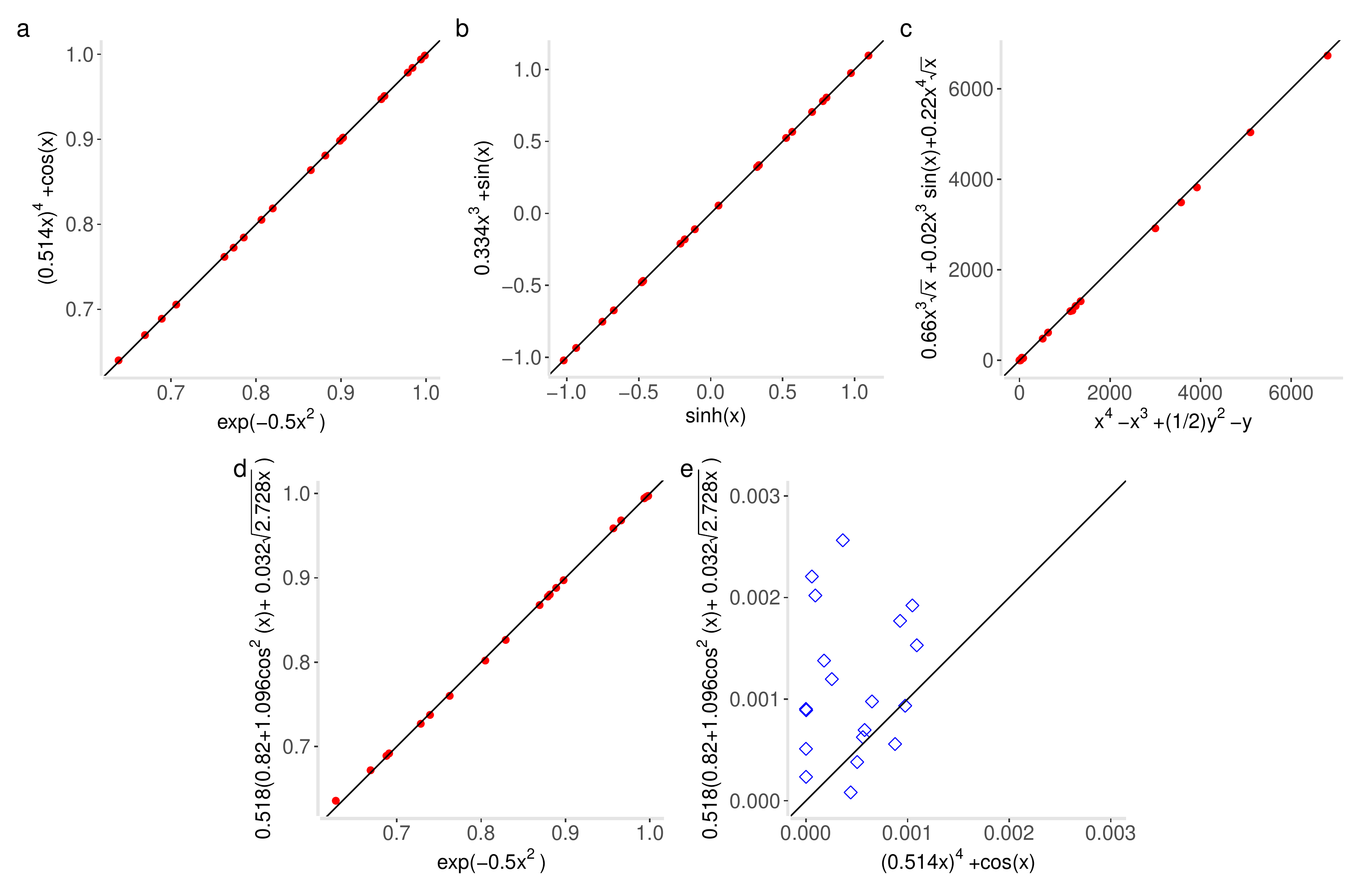}
\caption{\textbf{Concordance of benchmark and faiGP output.} \textbf{a)} Livermore-22: $\exp{(- 0.5x^2)} \approx (0.514x)^4+\cos{(x)}$ on $U(-1, 1, 20)$. The left-hand side is, up to the coefficient $1/\sqrt(2\pi)$, the probability density function of the standard normal distribution. \textbf{b)} Livermore-7: $\sinh{(x)} \approx 0.334x^3+\sin{(x)}$ on $U(-1, 1, 20)$. The (nonperiodic) hyperbolic sine is 'decomposed' as the sum of a (periodic) trigonometric function and a polynomial term (which thus quantifies departure from periodicity). \textbf{c)} Nguyen-12*: $ x^4 - x^3 + \frac{1}{2}y^2 - y \approx 0.66 x^3 \sqrt{x} + 0.02 x^3 \sin{(x)} + 0.22 x^4 \sqrt{x}$ on $U(0, 10, 20) \times U(0, 10, 20)$. The left-hand side is a function of two variables but is well-approximated by a function of a single variable. The original product domain $U(0, 10, 20) \times U(0, 10, 20)$ is projected onto the first component, on which the approximating function is defined. \textbf{d)} New Livermore-22: $\exp{(-0.5x^2)} \approx 0.518(0.82+1.096\cos^2{(x)}+0.032\sqrt{2.728x})$ on $U(-1, 1, 20)$. The right-hand side provides a new approximation of the standard normal density function using a trigonometric function, with maximal fitness score ($R^2=1$) as in panel \textbf{a}. \textbf{e)} The new Livermore-22 expression has a higher mean absolute error than the previous (panel \textbf{a}). Each point (a random draw from $U(-1, 1, 20)$) provides a comparison of the absolute error from the benchmark for the previous (x-axis) and new (y-axis) expression. The diagonal line with intercept at the origin is included in each panel to show departure from perfect concordance.}
\label{fig:comparisons_output}
\end{figure}

Some interesting patterns emerge (\textbf{Table~\ref{table:benchmarks}} and \textbf{Table~\ref{table:benchmarks_results}}). For 39 of the 43 benchmarks (90.7\%), faiGP recovers the generative function with Spearman correlation $\rho = 1$ ($R^2 = 1$) between the original and recovered datasets. The recovered expressions are exact for a number of functions; see, for example, the polynomial in ``Nguyen-1'' and $\sqrt{x}$ in ``Nguyen-8''. In most cases, faiGP generates a non-trivial equality or approximation to the original ground-truth expression (with maximum possible concordance, $R^2 = 1$). For example (see ``Livermore-22'' and ``Livermore-7'' and \textbf{Figure~\ref{fig:comparisons_output}a,b}): 

\begin{equation}
\exp{(- 0.5x^2)} \approx (0.514x)^4+\cos{(x)} \nonumber 
\end{equation}

 \noindent and 
 
\begin{equation}
\sinh{(x)} \approx 0.334x^3+\sin{(x)} \nonumber 
\end{equation}

 \noindent each defined on $U(-1, 1, 20)$ and having a mean absolute error of $1\text{e-}4$. The first relationship is interesting as the left-hand side defines, up to the coefficient $1/\sqrt(2\pi)$, the probability density function of the standard normal distribution. Note both the original and recovered functions are even functions, i.e., $f(x) = f(-x)$. Interestingly, the relationship connects a real-valued exponential function (which is not included in our library ${\cal F}$ or the training set) to a trigonometric function (a rare instance outside the well-known Euler's formula in the complex case) and selects the power (i.e., 4) of the polynomial term that is optimal among all integers in minimizing the mean absolute error of the recovered expression. The second relationship is equally interesting as an approximation to the hyperbolic sine (which is again not included in ${\cal F}$ or the training set). In this case, both the original and recovered functions are odd functions, i.e., $f(-x) = -f(x)$. Notably, the relationship suggests a connection between a nonperiodic hyperbolic function and a trigonometric function (and thus, a connection between the corresponding hyperbolic and Euclidean geometries and the obstruction to periodicity) without explicit use of complex numbers, akin to the Gudermannian function. In ``Nguyen-12*,'' the benchmark is a function of two variables but is well-approximated by a function of a single variable (\textbf{Figure~\ref{fig:comparisons_output}c}), as in dimensionality reduction.

 Note that faiGP performs poorly on ``Livermore-11'' and ``Livermore-12,'' which involve ratios of polynomials in two variables, but performs well on ratios of polynomials in a single variable (see ``R-1'' to ``R-3*''). The surfaces ``Livermore-11'' and ``Livermore-12'' \textit{blow up} when $x=y$ and $y=0$, respectively, within their respective domains whereas the ``R'' benchmarks do not have a singularity; hence, our default loss function ($\chi^2$ loss) is not as effective on the former benchmark set as on the latter.

\begin{figure}
\centering
\includegraphics[width=0.6\textwidth]{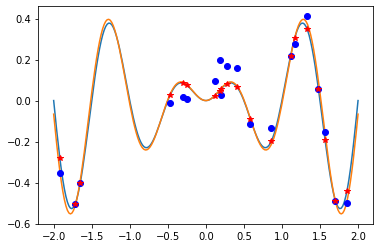}
\caption{\textbf{Impact of measurement noise.} We modeled noise as $\epsilon \sim \mathcal{N}(0, \lambda \times \sigma_f^{2})$ for a given benchmark function $f$. At a high level of noise, $\lambda = 0.072$, the faiGP output shows decrease in concordance with the original Keijzer-2 benchmark function (7.5\% exact recovery versus 100\% for $\lambda = 0.001$ [not shown]). Nevertheless, the approximating functional output (shown as blue curve) still shows high concordance. Blue dots are the noisy data ($f + \epsilon$) while red stars are the data points without noise ($f$).}
\refstepcounter{SIfig}
\label{fig:noise_lambda_064}
\end{figure}

For a given ground-truth expression $f$, we ask to what extent ``measurement noise'' can affect faiGP output. We consider the recovery performance in the presence of noise $\epsilon$: $f_{\textrm{with-noise}} = f + \epsilon$. We assume $\epsilon \sim \mathcal{N}(0, \lambda \times \sigma_f^{2})$, where $\sigma_f^{2}$ is the variance of $f(X)$ and $\lambda$ is a hyperparameter that determines the magnitude of the contribution of the error. Using ``Keijzer-2'' as the benchmark, at $\lambda = 0.001$, all runs recover expressions with high concordance ($R^2 \approx 1$) with a proportion (3 out of 10) recovering the exact expression. At higher level of noise, $\lambda = 0.072$, the average concordance for recovered expressions with the benchmark is reduced substantially although a proportion (2 out of 10) of the runs still recover the exact expression (\textbf{Figure~\ref{fig:noise_lambda_064}}). With even higher level of noise, $\lambda = 0.074$, none of the runs recover the exact expression and all show substantially degraded performance (i.e., maximum $R^2 \approx 0.80$).

Taken together, these results show that faiGP can generate some non-trivial symbolically equivalent expressions (``Ramanujan expressions'') or approximations with potentially interesting applications.

\par\null\par
\noindent \textbf{Hyperparameter tuning.}  We investigate the dependence of the fitness on hyperparameters. Recall that the smallest program is a 4-tuple consisting of a coefficient, an operator in the pre-selected library $\cal F$, an operand, and an exponent. Here, the accessible exponents during the evolutionary process are the integers within a pre-specified interval $[a, b]$, where $a$ and $b$ are hyperparameters. Some interesting observations must be noted. For the tested benchmarks, the $\chi^2$ loss shows improved accuracy for the symmetric exponent interval $a=-2, b=2$ relative to the non-symmetric interval $a=-3, b=2$. This is intuitively plausible as the non-symmetric interval makes a negative exponent more accessible to the loss function. For example, drawing the exponent from the interval $a=-3, b=2$, we obtain a new expression for the ``standard normal density'' benchmark (``Livermore-22''), which is now decomposed as a weighted sum of a trigonometric term (involving $\cos^2{(x)}$), as before, and the square root of (the absolute value of) $x$, with maximal concordance, $R^2=1$ (\textbf{Figure~\ref{fig:comparisons_output}d}):

\begin{equation}
\exp{(-0.5x^2)} \approx 0.518(0.82+1.096\cos^2{(x)}+0.032\sqrt{2.728x}) \nonumber 
\end{equation}

\noindent although the new expression is slightly less accurate, in terms of mean absolute error, than the previous one (\textbf{Figure~\ref{fig:comparisons_output}e}). On the other hand, ``Livermore-11'' and ``Livermore-12'' show substantial improvement in fitness score with these choices ($R^2 = 0.85$ and $R^2 = 0.98$, respectively), since these benchmarks already suffer from poor performance ($R^2 < 0$), under $\chi^2$ loss, from the presence of a singularity. Collectively, these results indicate that these specific hyperparameter choices may lead to considerable performance variation and some gain may be obtained by tuning them.

\begin{figure}
\centering
\includegraphics[width=1\textwidth]{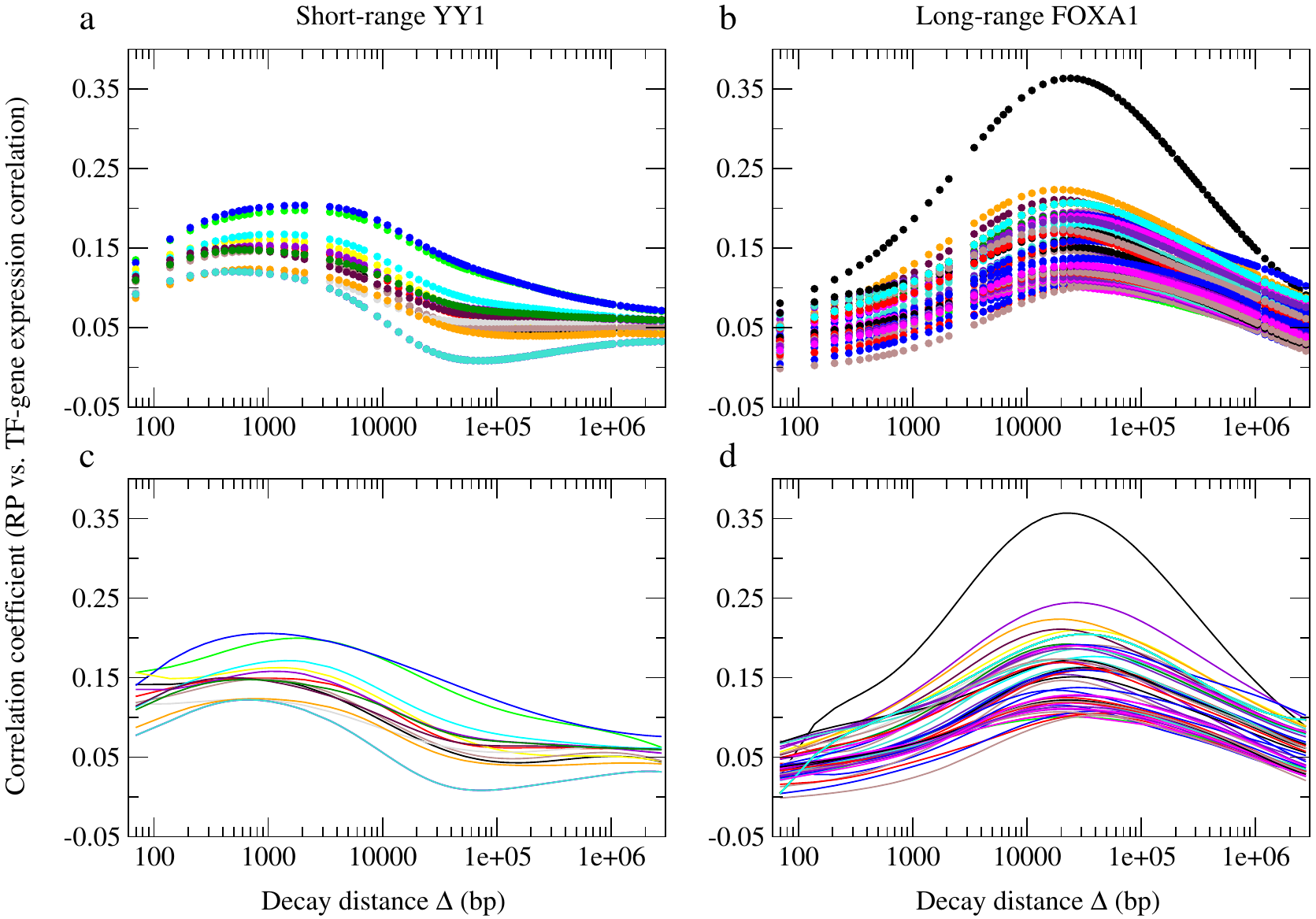}
\caption{\textbf{Modeling transcription factor regulatory range using faiGP.} For the transcription factor $i$ and target gene $j$, the agreement between the Regulatory Potential $RP_{ij}$ (which quantifies the sum of the regulatory effects of the binding sites of $i$ within the topologically associating domain [TAD] of $j$) and $\gamma_{ij}$ (which quantifies the effect of the perturbation of the expression of $i$ on the expression of the target $j$) is given by $\psi_i = \textrm{corr}(RP_{i*}, \gamma_{i*})$ (shown here on the y-axis). $RP_{ij}$ is a function of the decay distance $\Delta_i$ (shown here on the x-axis); thus, so is $\psi_i$. $\Delta^*_i$ is the point at which $\psi_i$ attains its maximum. We evaluate the performance of our framework by comparing the input dataset (top, each color representing a sample or condition from actual ChIP-seq and gene expression omics data \cite{chen2020determinants}) with the corresponding output (bottom) from the faiGP-derived mathematical expression. In the columns, we present a transcription factor with a short regulatory range (YY1) and another with a long regulatory range (FOXA1). \textbf{a)} Short-range $\Delta^*_i
~(100 ~ \textrm{bp} -3 ~\textrm{kb})$ TF - YY1 \textbf{b)} Long-range $\Delta^*_i
~(3 ~ \textrm{kb} -100 ~\textrm{kb})$ TF - FOXA1 \textbf{c)} Results from recovered expression for YY1 \textbf{d)} Results from recovered expression for FOXA1.}
\label{fig:TF}
\end{figure}

\par\null\par

\noindent \textbf{Applications.} Here, we demonstrate how the framework can be used in scientific modeling from the mathematics of reaction networks (kinetics) and, in a related problem, a model of transcription factor regulation of gene expression.
\subsubsection*{Ligand-receptor binding kinetics and gene transcription regulation}
A number of models have been developed to characterize the binding of transcription factors (TFs) to their regulatory DNA target sites (enhancers and silencers) to regulate gene transcription. The models come from such diverse approaches as differential equations \cite{narang2006comparative}, statistical physics \cite{chu2009models}, and mass action kinetics (Hill-Langmuir equation) \cite{santillan2008use}. Here we explore a potential application of our neural-network-guided genetic programmer by considering the faiGP-derived Ramanujan expression in relation to these models.

Hill functions describe the equilibrium state of the \textit{simultaneous} binding of multiple ligands to a target molecule. The functions can be used to model the rate of gene transcription via the regulation of TFs at (multiple) binding sites. Let us assume $n$ molecules of the ligand $T$ for the receptor $D$. The reaction of interest to us is the following:

\begin{align}
D + nT \ce{<=>[\ce{k_f}][\ce{k_b}]} D_{nT} \nonumber \\ 
[D][T]^n = K[D_{nT}] \label{eq:hilleqn}
\end{align}
       
\noindent where the notation $[A]$ denotes the concentration of the molecule $A$, $k_f$ and $k_b$ are the on-rate and off-rate constants, respectively, and $K \coloneqq \frac{k_b}{k_f}$ is the dissociation constant. The dissociation constant $K$ can also be calculated from the thermodynamic equation:

\begin{equation}
K = e^{(-\frac{\Delta G_0}{R \tau})} \nonumber
\end{equation} 

\noindent where $\tau$ is the absolute temperature, $R$ is the universal gas constant, and $\Delta G_0$ is the Gibbs free energy change. Assuming a fixed number of receptors, the proportion of bound receptors, i.e., the probability of binding event, as a function of $[T]$ can be derived from equation (\ref{eq:hilleqn}):

\begin{equation}
    p_{\mathrm{bound}}([T]) = \frac{[T]^n}{K + [T]^n} = \frac {1} {1 + (\frac{K}{[T]^n})} = \frac {1} {1 + (\frac{K_{{1/2}}}{[T]})^n}
    \label{eq:pbound}
\end{equation}

\noindent where $K_{{1/2}}$ is the ligand concentration that produces half occupancy. Note $p_{\mathrm{bound}}([T])$ can be used to model the probability that a gene is transcribed as a function of the concentration $[T]$. Indeed, the dynamics of the transcription of the gene $G$ regulated by the transcription factor $T$ can be modeled as follows:

\begin{equation}
\frac{d[G]}{dt} = \gamma_G [D] p_{\mathrm{bound}}([T])
\end{equation}

\noindent where $\gamma_G$ is the rate of transcription. (An additional RNA decay term may be included.) 

\begin{figure}
\centering
\includegraphics[trim=0cm 0cm 0cm 0cm,clip=true,width=1\textwidth,height=0.3\textwidth]{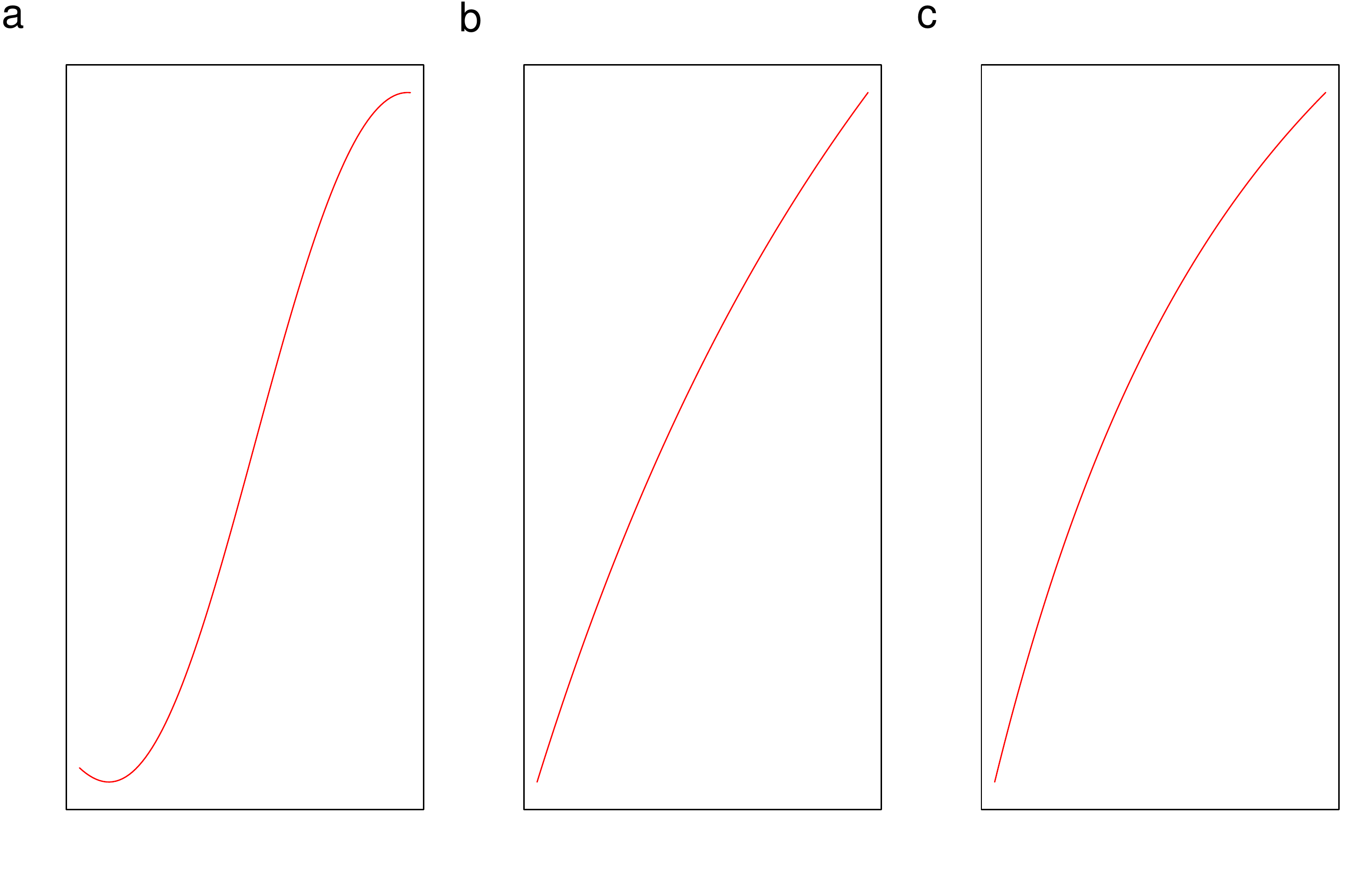}
\caption{\textbf{Sigmoidal behavior of faiGP output for Hill functions.} \textbf{a)} faiGP output ($C x^2 \cos^2 (\omega \log(\nu x))$), defined on an interval $(a,b)$, $b > a$, in comparison with Hill function with \textbf{b)} Hill coefficient $n=1$ and \textbf{c)} Hill coefficient $n=2$. The input comes from an experimental dataset \cite{gadagkar2015computational}. Here, the range for the x-axis is the interval $(10, 16)$, $C = 0.4463487$, $\omega = -4.333$, and $\nu = 1.201$. An arbitrary interval $(c,d)$ can be shifted onto this interval $(a,b)$ via a shift transformation $s(x) = a + (\frac{b-a}{d-c})(x-c)$.}
\refstepcounter{SIfig}
\label{fig:candidate_hill}
\end{figure}

However, the assumption of simultaneous binding to a target is a physically unrealistic reaction condition characterized by extreme cooperativity among the binding sites \cite{weiss1997hill}. Thus, a symbolic expression for $p_{\mathrm{bound}}([T])$ (derived from experimental data) equivalent to equation (\ref{eq:pbound}) can have broad applications. We apply our framework to dose-response experimental data \cite{gadagkar2015computational} that have been used to fit the Hill equation and obtain the following expression, defined on an interval $(a,b)$ with $b > a$, with high sigmoidal behavior (\textbf{Figure~\ref{fig:candidate_hill}}):

\begin{align}
p_{\mathrm{bound}}([T]) &= C [T]^2 \cos^2 (\omega \log(\nu [T])) \nonumber \\
&= C [T]^2 \cos^2 (\omega \log(\nu) + \omega \log([T])) \nonumber \\
&= C [T]^2 \cos^2 (\theta + \omega \log([T]))
\end{align}

\noindent  Here, $\omega$ and $\nu$ are nonzero constants (which control the degree of sigmoidicity), $\theta = \omega \log(\nu)$, and $C = \nu^2p_{\mathrm{bound}}(\frac{1}{\nu})$, as can be seen by setting $[T] = \frac{1}{\nu}$. The change of variables $[T] = e^u$ linearizes the argument of the cosine operator:

\begin{equation}
p_{\mathrm{bound}}([T]) = r_{\mathrm{bound}}(u) = C e^{2u} \cos^2 (\theta + \omega u) = \frac{1}{2} C e^{2u} [1 +  \cos(2(\theta + \omega u))]
\label{eq:faiGP_LR_model}
\end{equation}

\noindent The last equality in equation (\ref{eq:faiGP_LR_model}) follows from the trigonometric identity for $\cos^2(x)$.

\begin{figure}
\centering
\includegraphics[trim=0cm 0cm 0cm 0cm,clip=true,width=0.8\textwidth,height=0.6\textwidth]{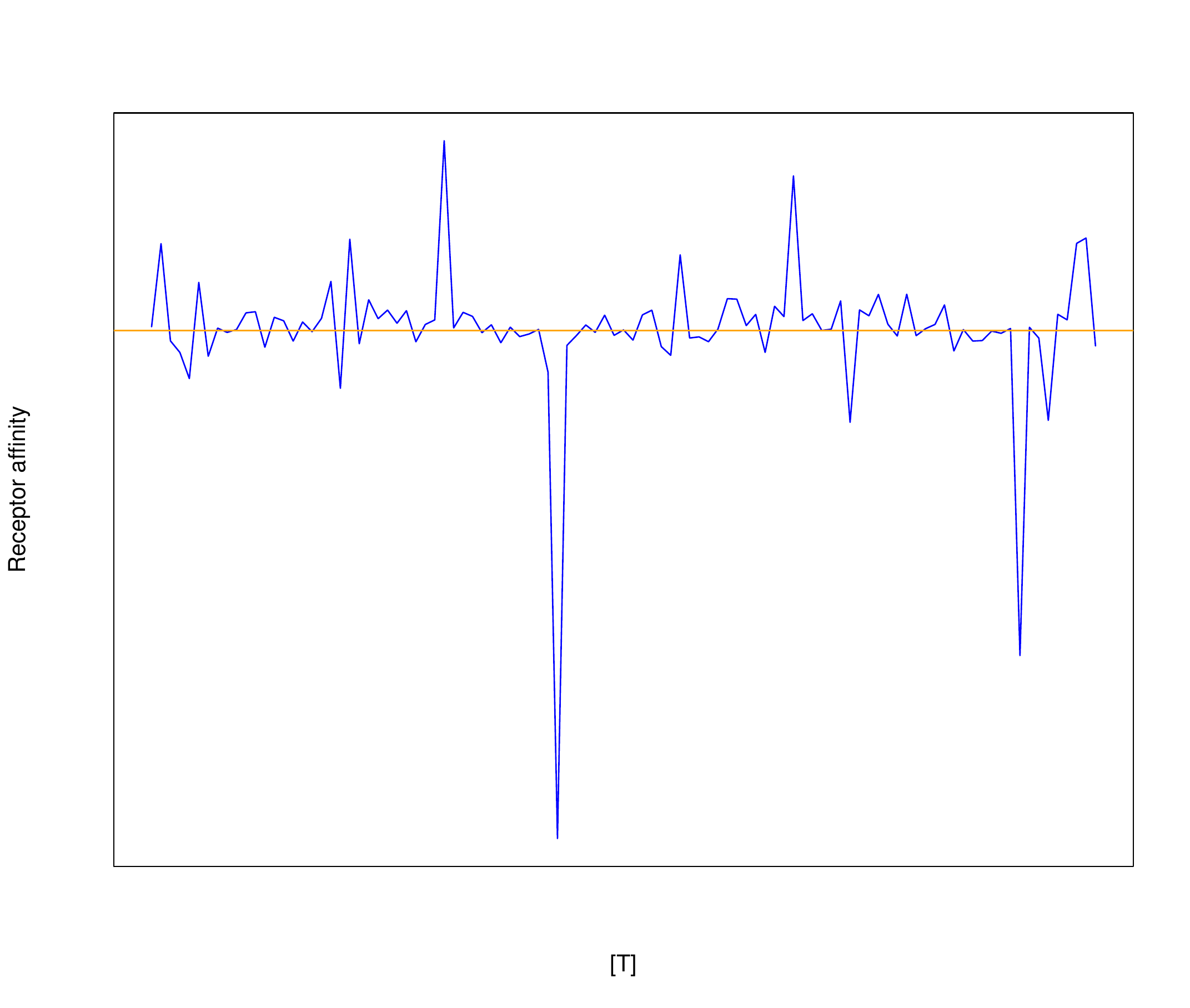}
\caption{\textbf{Cooperativity index as a function of $[T]$.} From the faiGP formulation and in contrast to the conventional Hill equation, the impact of ligand binding on the receptor's affinity for additional ligand binding depends on ligand concentration and may undergo ``alternation'' of positive cooperativity and negative cooperativity. Orange line represents $n_H = 1$. For this example, the input is from experimental Hill dose-response data \cite{gadagkar2015computational}. Here, the range for the x-axis is the interval $(10, 16)$, $C = 0.4463487$, $\omega = -4.333$, and $\nu = 1.201$ for the faiGP output ($x \rightarrow C x^2 \cos^2 (\omega \log(\nu x))$); see \textbf{Figure~\ref{fig:candidate_hill}a} for the corresponding plot of $p_{\mathrm{bound}}([T])$.}
\refstepcounter{SIfig}
\label{fig:receptor_affinity}
\end{figure}

We define the ``cooperativity index'': 

\begin{equation}
n_H \coloneqq \frac {d \log (\frac {p_{bound}([T])}{1-p_{bound}([T])})}{d(\log([T]))}
\end{equation}

\noindent as a measure of how fast the ``odds ratio'' of the binding changes with respect to the concentration $[T]$ (in log scale). As a biochemical metric, $n_H$ gives an indication of the impact of ligand binding on the receptor's apparent affinity for another ligand molecule binding. From equation (\ref{eq:pbound}), the cooperativity index equals the Hill coefficient of the biochemical reaction, i.e., $n_H = n$, under the conditions in which the Hill equation is applied. Thus, under these assumptions, cooperativity is fixed, with the receptor showing a constant affinity for another ligand binding. However, since the Hill formulation is reliable only in the extreme case of simultaneous binding, we calculate the cooperativity index from the faiGP-derived formalism (equation (\ref{eq:faiGP_LR_model})). We find that, in contrast to a constant value under the unrealistic conditions of the Hill equation, the effect of ligand binding on the receptor's affinity for additional ligand binding can be highly variable and may attain local maxima (where positive cooperativity, i.e., $n_H > 1$, is also observed) and local minima (where negative cooperativity, i.e., $n_h < 1$, is also observed), depending on the ligand concentration $[T]$ (\textbf{Figure~\ref{fig:receptor_affinity}}).

\subsubsection*{Regulatory range of the cistrome}
A related problem is to quantify the genomic distance over which transcription factors regulate gene expression. To extend the ligand-receptor interaction kinetics model with its specific application to gene transcription, we consider the ``Regulatory Potential''(RP) \cite{chen2020determinants}, which models the contribution of the transcription factor $i$ on the target gene $j$ as a sum of regulatory effects of multiple binding sites near the transcription start site (TSS) of $j$ within the topologically associating domain (TAD) of $j$. From Chromatin Immunoprecipitation Sequencing (ChIP-seq) data, the regulatory effect of a ``peak'' $k$ for $i$ on the gene $j$ is modeled using the genomic distance $d_{ijk}$ from the TSS of $j$ to $k$. RP is parameterized by the ``decay distance'' $\Delta_i$ as follows:

\begin{equation}
RP_{ij}(\Delta_i) = \sum_{\textrm{peak} k} 2^ \frac{-d_{ijk}}{\Delta_i}
\label{eq:RP}
\end{equation}

\noindent Now consider the correlation $\gamma_{ij} = \textrm{corr}(E_i, E_j)$ of the expression of $i$ and $j$. The agreement between the regulatory potential $RP_{i*}$ (defined to model the regulatory effects of the cistrome) and $\gamma_{i*}$ (defined as a proxy of the effect of transcription factor perturbation on target gene expression) can be quantified from the correlation:

\begin{equation}
\psi_i = \textrm{corr}(RP_{i*}, \gamma_{i*})
\label{eq:RP_cor_agree}
\end{equation}

\begin{figure}
\centering
\includegraphics[width=0.8\textwidth]{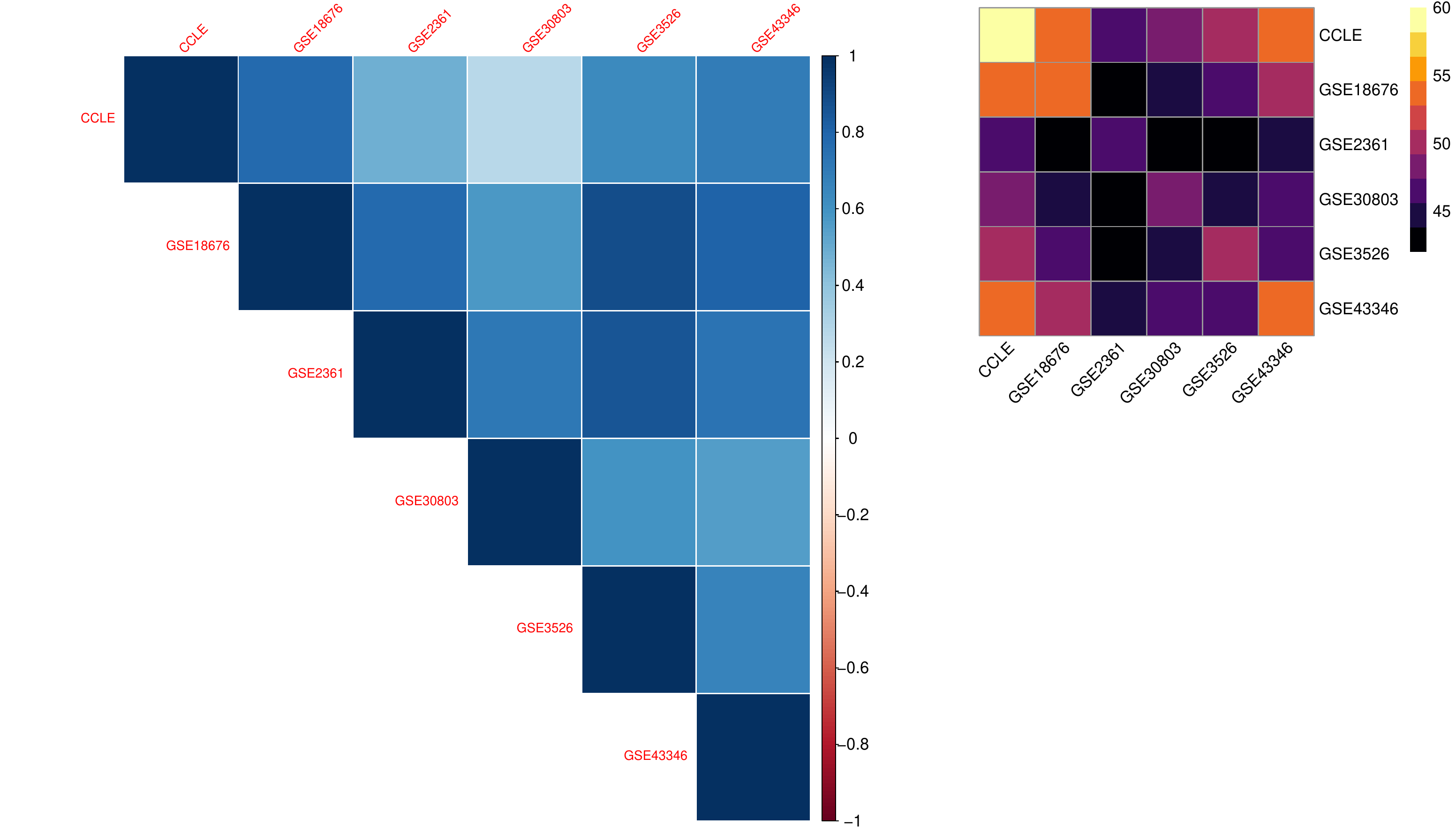}
\caption{\textbf{Correlation of regulatory decay distance across tissue datasets.} Let $\Gamma$ be the matrix of regulatory decay distance values $\Delta^*_i$, where the rows correspond to the transcription factors and the columns the tissue datasets. The $lm$-th entry of the correlation matrix (left panel) equals the correlation between tissue datasets $l$ and $m$. Let $\Gamma_0$ be the corresponding missing-value indicator matrix (of the same matrix dimension as $\Gamma$), where each cell is boolean, i.e., $0$ indicates missing value and $1$ otherwise. The $lm$-th entry of the product matrix $\Gamma^T_0 \Gamma_0$ is the total number of paired non-missing values for the tissue datasets $l$ and $m$ (right panel). On our reanalysis of data from Chen et al. \cite{chen2020determinants}, substantial pairwise correlation of the $\Delta^*_i$ between tissue datasets across the 60 transcription factors (11 short-range and 49 long-range) is observed (left panel). We include only datasets with at least half of the transcription factors having non-missing values.}
\refstepcounter{SIfig}
\label{fig:Delta_star_correlation_across_TF_between_tissues}
\end{figure}

\noindent Note that $\psi_i$ is a function of the decay distance $\Delta_i$. The regulatory decay distance $\Delta^*_i$ of the transcription factor $i$ is defined as the optimal decay distance that maximally separates the differentially expressed genes, upon perturbation of the expression of the transcription factor, from the remaining genes. That is, $\Delta^*_i$ can be estimated by the decay distance at which $\psi_i$ attains a maximum. Across the transcription factors evaluated (11 short-range and 49 long-range), the parameter $\Delta^*_i$ shows high correlation between tissue expression datasets (\textbf{Figure~\ref{fig:Delta_star_correlation_across_TF_between_tissues}}). (See Chen et al. \cite{chen2020determinants} for additional details. See also \textbf{Supplementary Note} for a discussion of the caveats that arise from a noisy estimate $\widehat{\gamma_{ij}}$ of the true $\gamma_{ij}$ and a presentation of a simulation framework to study the theoretical properties of $\psi_i$.)

Note that $\gamma_{ij}$ equals the effect of the transcription factor $i$ on the gene $j$ under a linear model with (zero-mean) Gaussian total contribution from additional determinants. More generally, the quantity of interest to us is $\frac{\partial g}{\partial E_i}$ from the (generally unknown and possibly nonlinear) expression $E_j = g(E_i)$; assuming standardized expression levels, then, in the specific case of the linear model, $\frac{\partial g}{\partial E_i} = \gamma_{ij}$. Notably, the function $g$ can be modeled as a function of the level of concentration of $i$ using faiGP-based $p_{\mathrm{bound}}(E_i)$ (equation (\ref{eq:faiGP_LR_model})). In the formulation above (equation (\ref{eq:RP})), the regulatory influence of a single peak at distance $k$ to the TSS of the gene $j$ is modeled as an exponential decay function, with $\Delta_i$ defining the the half-life. Again, more generally, using matched ChIP-seq and tissue-specific gene expression data \cite{mei2016cistrome, gamazon2018using}, a symbolic expression $h_i(d)$ of the regulatory influence as a function of the distance $d$ may be derived via faiGP; this function is potentially transcription-factor-specific.

We apply our framework to discover a symbolic expression for $\psi_i(\Delta_i)$ (equation (\ref{eq:RP_cor_agree})). As a measure of the agreement between the regulatory effects of transcription factor binding sites and the perturbation effect of a transcription factor on its target,  $\psi_i$ integrates cross-omics (ChIP-seq and gene expression) data. We find that the framework can derive a mathematical expression for $\psi_i$ as a function of the decay distance $\Delta_i$, which shows high concordance with the original input data (\textbf{Figure~\ref{fig:TF}}). (See also our simulation results above in the presence of ``measurement noise'' for additional context.) In particular, this expression provides a closed-form for $\frac{\partial  \psi_i}{\partial \Delta_i}$ (as well as the second derivative) and, thus, a principled approach to derive the optimal $\Delta^*_i$ and study the global (genome-wide) behavior of $\psi_i$.

\subsubsection*{Transcription-factor-specific unified expression characterizing the regulatory range of the cistrome}
As defined, $\psi_i$, as a function of the decay distance $\Delta_i$, is actually parameterized by a variable $s$ denoting a sample (cell type or condition) (\textbf{Figure~\ref{fig:TF}a and b}). Thus, we search for a function $\Psi_i$ that reflects this dependence on both $\Delta_i$ and $s$ such that the original $\psi_i$ can be recovered by setting $s \in \mathbb{R}$:

\begin{equation}
\Psi_i(\Delta_i, s) = \psi^s_i(\Delta_i) \nonumber
\end{equation}

\begin{figure}
\centering
\includegraphics[width=0.9\textwidth]{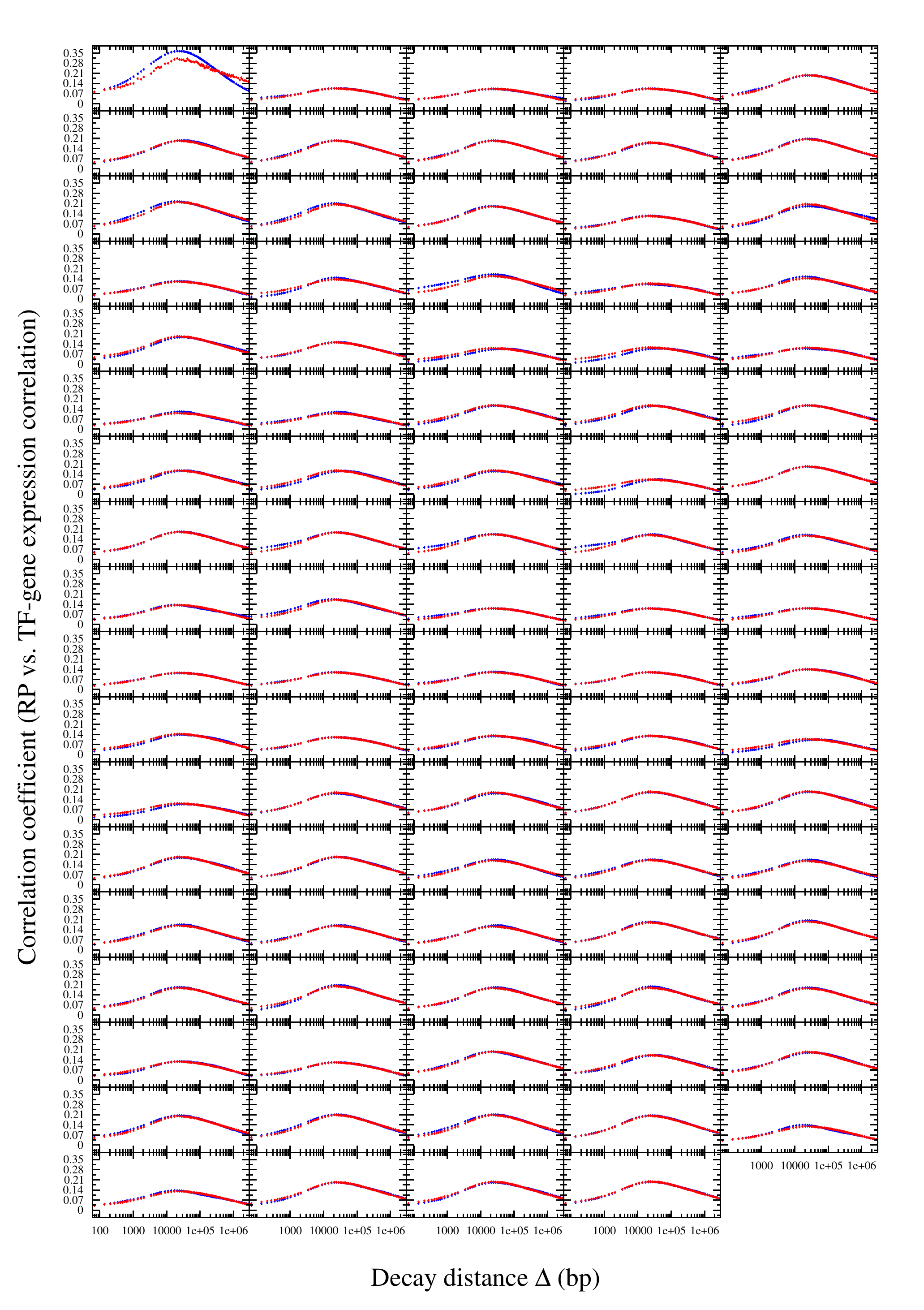}
\caption{FOXA1 data (blue dots) in comparison with the autoencoder output (red dots).}
\refstepcounter{SIfig}
\label{fig:FOXA1_AE}
\end{figure}

\begin{figure}
\centering
\includegraphics[angle=-90,width=1\textwidth]{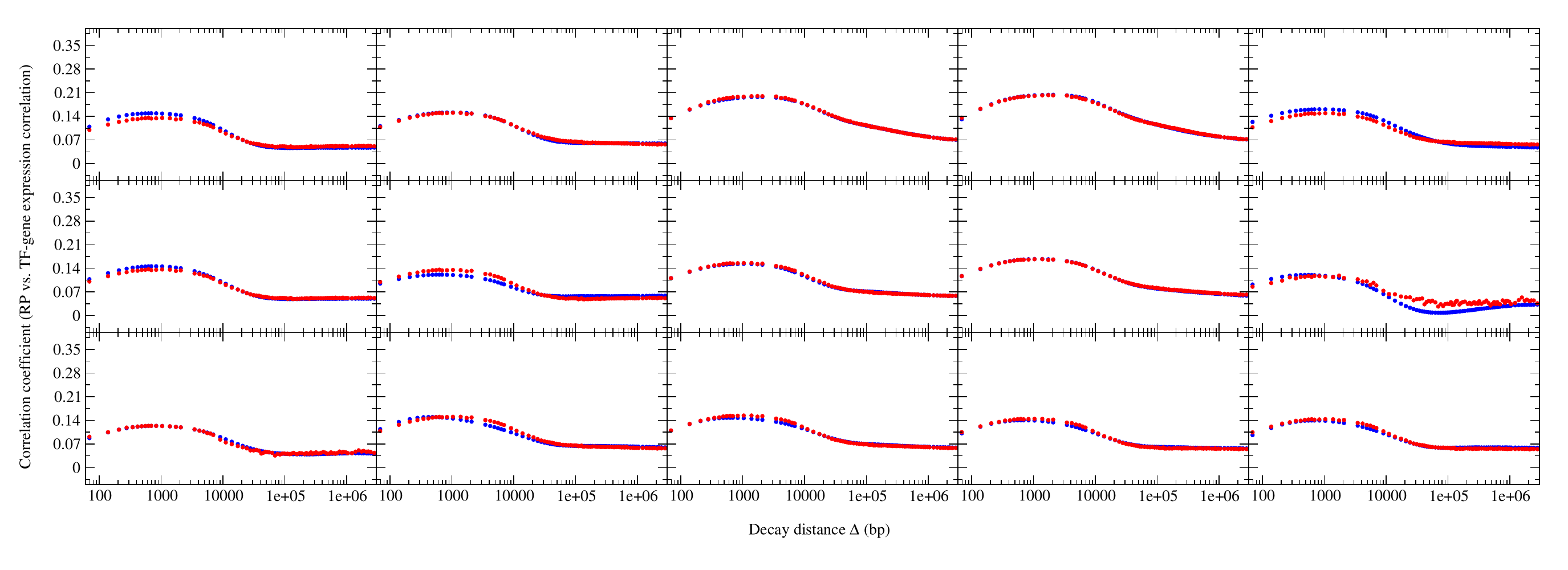}
\caption{YY1 data (blue dots) in comparison with the autoencoder output (red dots).}
\refstepcounter{SIfig}
\label{fig:YY1_AE}
\end{figure}

\noindent Here, we use $\psi^s_i$ to emphasize the dependence of the original $\psi_i$ on $s$. We can think of $\Psi_i$ as an extension of $\psi_i$ to a higher-dimensional domain. Geometrically, we can view each curve $\psi_i$ as the ``isocurve'' of the surface $\Psi_i$ at some $s$. Note the parameterization is not uniquely determined; for example, if $s$ is a valid parameterization, so is $\tau(s)$, for any monotonically increasing function $\tau$. We determine the value of $s$ for each sample using an autoencoder, trained on 90\% of the data, with a one-dimensional latent space (see \textbf{Figure~\ref{fig:FOXA1_AE}} and \textbf{Figure~\ref{fig:YY1_AE}}, which show autoencoder reconstruction concordance with the original data from the latent-space parameterization $s$). Using faiGP, we find such a function $\Psi_i$ for FOXA1 with $R^2 = 0.95$ and for YY1 with $R^2 = 0.95$, where the $R^2$ is calculated from the test data (30\% of the samples):

\begin{align}
&\Psi_{\textrm{FOXA1}}(\delta \coloneqq \log{(\Delta_\textrm{FOXA1})}, s) = 0.021  \sqrt{s} ~ (7.13 + 3.402 \cos{(0.607 \delta)}) \nonumber \\
&\Psi_{\textrm{YY1}}(\delta \coloneqq \log{(\Delta_\textrm{YY1})}, s) = 0.057(-1.52 s - 0.47 \delta + 0.372 \cos{(0.853 \delta)}+2.474 \log{(0.987 \delta)}) \nonumber
\end{align}

\begin{figure}
\centering
\includegraphics[width=0.9\textwidth]{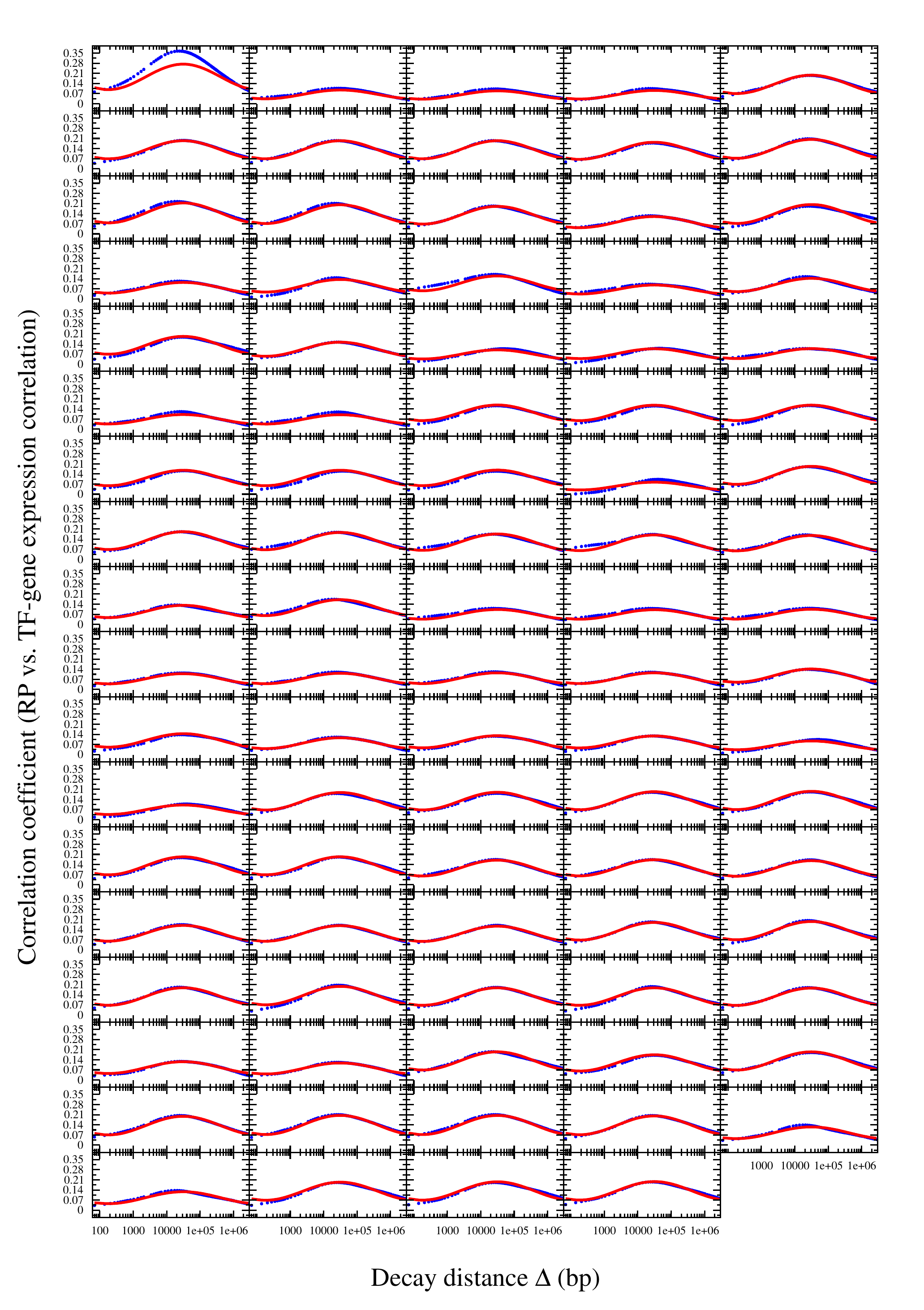}
\caption{FOXA1 data (blue dots) in comparison with the faiGP output (red curve) for $\Psi_{\textrm{FOXA1}}$ at distinct values of $s$.}
\refstepcounter{SIfig}
\label{fig:FOXA1_faiGP}
\end{figure}

\begin{figure}
\centering
\includegraphics[angle=-90,width=1\textwidth]{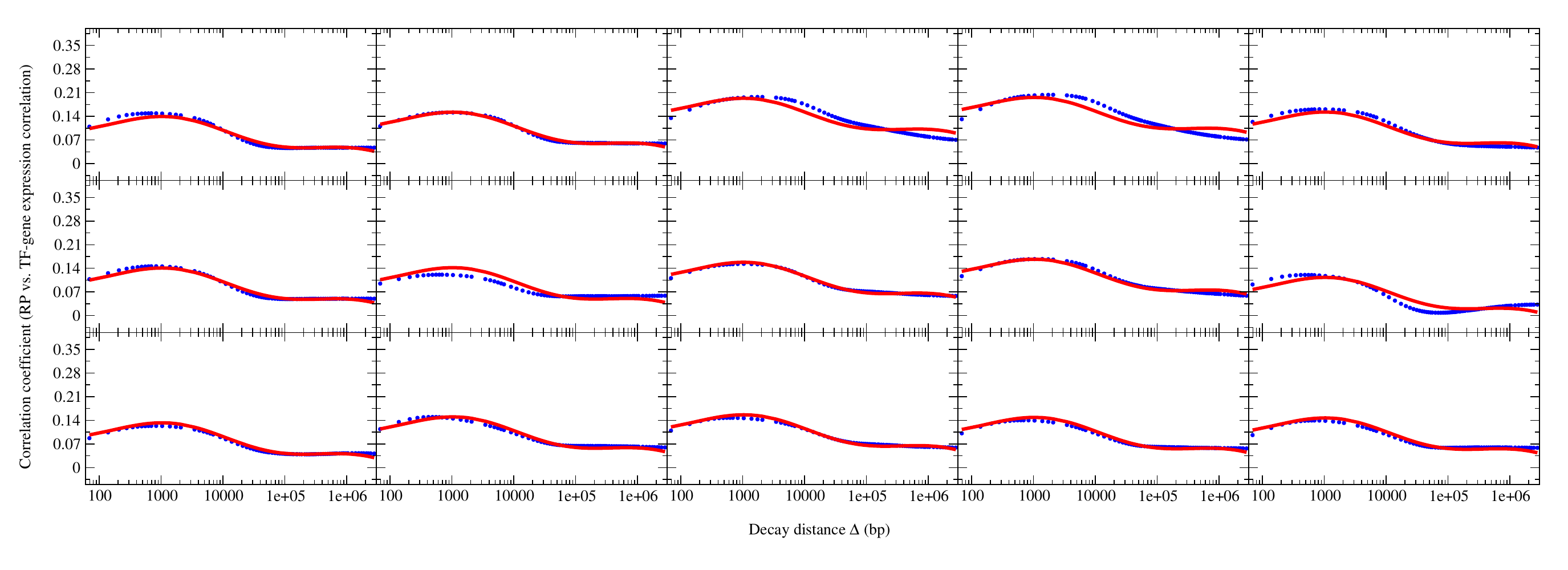}
\caption{YY1 data (blue dots) in comparison with the faiGP output (red curve) for $\Psi_{\textrm{YY1}}$ at distinct values of $s$.}
\refstepcounter{SIfig}
\label{fig:YY1_faiGP}
\end{figure}

\noindent (See \textbf{Figure~\ref{fig:FOXA1_faiGP}} and \textbf{Figure~\ref{fig:YY1_faiGP}} for comparison of the original data with the faiGP output $\Psi_i$.) In both cases, we obtain the expression:  

\begin{equation}
\Psi_i (\delta, s) = A(s) \cos{(B \delta)} + C(\delta, s)
\end{equation}

\noindent where the amplitude $A(s)$ is a function of $s$, $\frac{2 \pi}{B}$ is the (constant) period for a positive real number $B \in \mathbb{R}^+$, and the vertical shift $C(\delta, s)$ is a function of the decay distance and the sample. For example, the amplitude and vertical shift for FOXA1 are determined by $\sqrt{s}$ (and independent of $\delta$) whereas the period and frequency can be obtained (i.e., $\frac{2 \pi}{0.607}$ and $\frac{0.607}{2 \pi}$, respectively) from the constant coefficient of $\delta$. Note the concordance for $\psi_i$ (\textbf{Figure~\ref{fig:TF}}) is slightly greater than that for $\Psi_i$ (see the discordance in the first cell in \textbf{Figure~\ref{fig:FOXA1_faiGP}}), suggesting that a second parameter $t$ (specific to the sample) is needed to close the gap. In summary, a high-fitness equation for $\Psi_i$ is generated which, in effect, stitches together the various condition-dependent ($s$-parameterized) functions $\psi_i$ and which may be further enhanced by extension of the domain to a higher dimension, i.e., inclusion of additional parameters.

\section*{Discussion} \label{sec:SUM}

Here, we develop an integrative approach incorporating a convolutional variational autoencoder (for dimensionality reduction to reduce the search space), a bayesian multilabel classifier (for posterior inference and generation of the prior to guide the search), and a genetic programmer (for evolution of programs towards optimal fitness), to perform symbolic regression. A new grammar exploits the features of an algebra that provides a universal approximation property and minimizes bloat, in order to generate syntactically valid and well-defined expressions. By design, faiGP is extensible; for example, the grammar can be generalized to encode the $\mathbb{C}$-algebra of complex-valued functions.

We explore the impact of design choices, including the loss function, the regularizer (for diversity and complexity), and coefficient assignment. Evaluation of benchmarks highlights the method's strengths, including the generation of interesting Ramanujan expressions, and limitations, notably, the challenge of recovery still for certain benchmarks. We find that using ConVAE and BMC to inform the genetic programming leads to improvement in the recovery of the ground-truth expression. In future work, given the observation that hyperparameters may produce performance variation, a neural network may be designed to learn optimized values based on specific applications. 

We demonstrate an application of the framework in a theoretical account of ligand-receptor interactions with immediate relevance to transcription factor binding to regulatory DNA sequence. We propose a faiGP-derived model of ligand-receptor binding kinetics that is generated from Hill dose-response experimental observations, providing a transcription-factor-mediated model of gene expression. Finally, the development of a model of transcription factor regulatory range with high fidelity to the original data, as an application, suggests the framework can facilitate discovery of governing equations in high-dimensional genomic data.

\begin{acknowledgments}
This research was supported by National Institutes of Health (NIH) grants NHGRI R35HG010718, NHGRI R01HG011138, NIA AG068026, and NIGMS R01GM140287 to E.R.G..
\end{acknowledgments}

\bibliography{bibliography}
\newpage

\beginsupplement
\begin{center}
\textbf{\large Supplementary Materials: Neural-Network-Directed Discovery of Governing Equations using Function-Algebra-Informed Genetic Programming}
\end{center}
\newpage

\section*{Supplementary Note}
\noindent \textbf{An overview of the evolutionary processes.} The evolutionary processes in faiGP are defined similarly to those of conventional GP with some customizations so as to be compatible with the new grammar. Here, we give a comprehensive overview of these processes:

\begin{itemize}
\item Point mutation: Each element of a 4-tuple program can be mutated under point mutation as long as the resulting program is still a syntactically valid program, i.e., $C \rightarrow  C'$, $F \rightarrow F'$, $O \rightarrow O'$, and $P \rightarrow P'$ where the primed elements are drawn from the same set as their unprimed counterparts. The probability by which each element is chosen for mutation is as follows: the coefficient $C$ and the operand $O$ with probability 15\% each, the operator $F$ with probability 82.5\%, and the exponent $P$ with probability 2.5\%.

\item Crossover: During a crossover operation, the selection of the segments from the parent programs is executed at random and the resulting program is validated according to the grammar; a similar process happens in conventional GP. Consistent with Rule 6 of the grammar, operands and programs are exchangeable during crossover, i.e., $O \rightarrow \{(C, F, O, P)\}$ and vice versa. With the exception of this allowed exchange, selected parental segments in a crossover should be equivalent, i.e., contain elements of the same type, e.g., $\{(F, O)\}$ and $\{(F', O')\}$.

\item Subtree mutation: A subtree mutation is a crossover operation in which one of the parents is not chosen from the pool of programs in that generation. Instead, that parent is generated according to the prior probabilities given to faiGP by BMC.

\item Hoist mutation: In hoist mutation, a segment of a program is selected to be replaced. A smaller segment of the selected segment is then selected and is ``hoisted" in the place of the original segment.

\item Replication: An exact copy of the original program is generated.
\end{itemize}

\par\null\par

\begin{figure}
    \centering
    \includegraphics[trim=0cm 0cm 0cm 0cm,clip=true,width=0.7\textwidth]{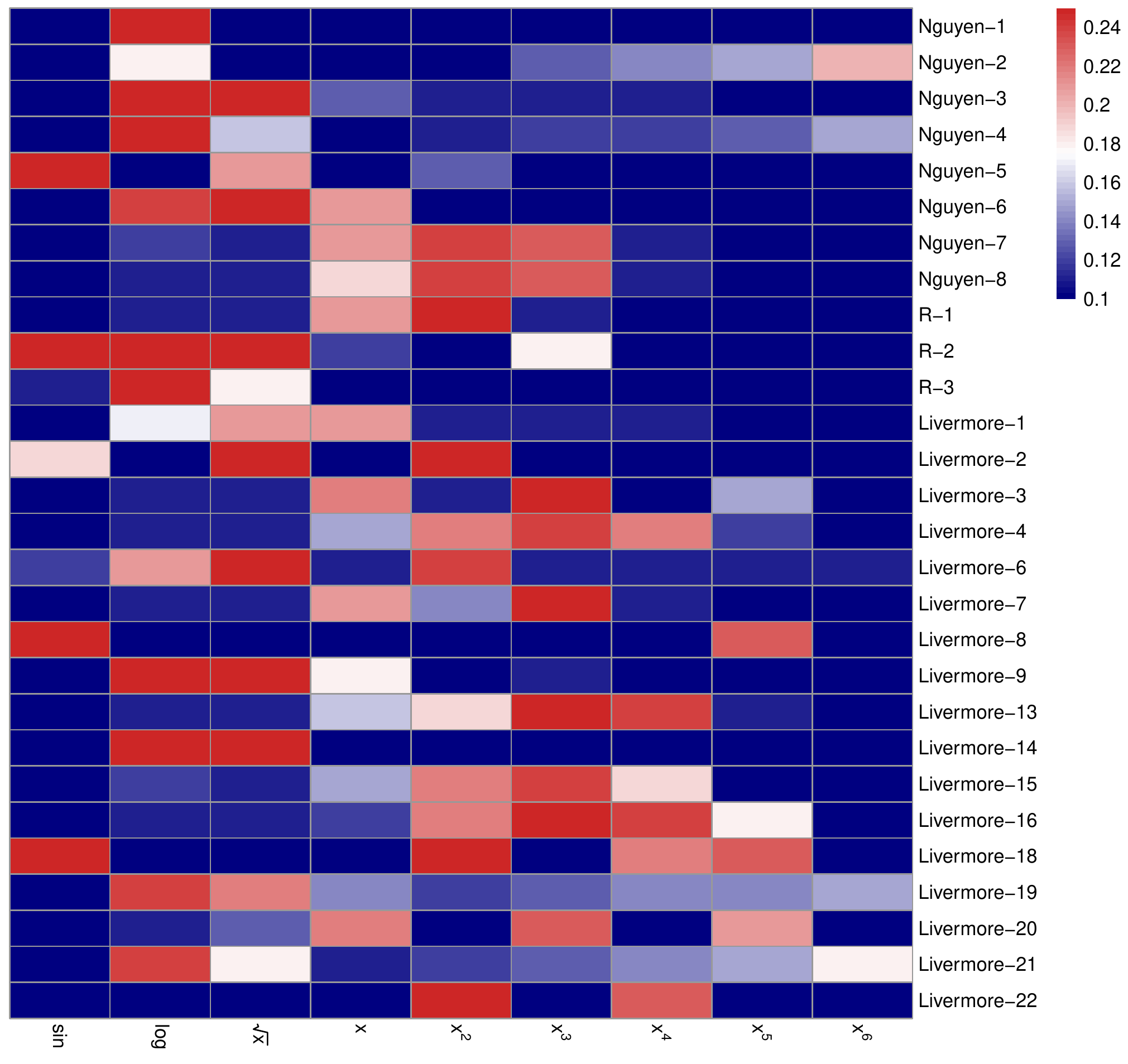}
    \caption{\textbf{Posterior probabilities for benchmark functions.} For each benchmark function (row), the posterior probability $p(O | \bm{z}, {\cal D})$ of an operator $O$ (column) given the latent space $\bm{z}$ and data ${\cal D}$ is generated from the application of ConVAE and BMC. We note several patterns, some quite intuitive and others less so. Based on the posteriors, across these benchmarks, the high-degree monomial $x^6$ will be less frequently used to inform the faiGP search than the low-degree monomial $x^2$. On ``Nguyen-5'' (i.e., $\sin{(x^2)}\cos{(x)}-1$), the sine operator obtains the highest posterior and will therefore initiate the faiGP search. On ``Livermore-8'' (i.e., $\cosh{(x)}$), which is not part of the library $\cal F$ of operators of the faiGP grammar, the sine operator, which has the highest posterior, will be used by faiGP. On ``Livermore-3'' (i.e., $\sin{(x^3)}\cos{(x^2)}-1$), the function $x^3$ stands out for its posterior and will then be used to initialize faiGP. ``Livermore-22'' (i.e., $\exp{(- 0.5x^2)}$), which is related to the standard normal density function, will be initiated by $x^2$ based on its posterior; we note the exponential operator is not part of the library $\cal F$. Less intuitively, $x$, $x^2$, and $x^3$ are assigned higher posteriors than $\sqrt{x}$ for the exact benchmark $\sqrt{x}$ of ``Nguyen-8''. The $\log$ and $\sqrt{x}$ operators are relatively highly correlated (Spearman's $\rho = 0.50$) across these benchmarks. Driven by the relatively low posterior probabilities across the benchmarks, the correlation (Spearman's $\rho = 0.42$) between $x^5$ and $x^6$ is relatively high.}
    \refstepcounter{SIfig}
    \label{fig:bmc_output}
\end{figure}

\noindent \textbf{Posteriors of $\log$ for the benchmark functions.} The (natural) $\log$ operator obtains a high posterior probability (equation (\ref{eq:posterior})) for many benchmark functions among the operators in the library $\cal F$ (\textbf{Figure~\ref{fig:bmc_output}}). Here, we provide a possible explanation. Let $y(n) = 10^n$ be defined on the set of integers. Any function $f : K \rightarrow \mathbb{R}$ can be represented as a set of ordered 2-tuples $\{(x, f(x))\}$. The functional value $f(x) \in \mathbb{R}$ at $x$ can be written in ``decimal notation'' as follows:

\begin{equation}
f(x) = a_{n(x)}(x) 10^{n(x)} + a_{n(x)-1}(x) 10^{n(x)-1} + \ldots + a_1(x) 10^1 + a_0(x) + a_{-1}(x) 10^{-1} + \ldots + a_{-m}(x) 10^{-m} \nonumber 
\end{equation}

\noindent where $m$ determines the desired number of significant digits to declare equality and is fixed. This expression can be rewritten (using the definition of $y(n)$):

\begin{equation}
f(x) = a_{n(x)}(x) 10^{\log_{10}y(n(x))} + a_{n(x)-1}(x) 10^{\log_{10}y(n(x)-1)} + \ldots + a_1(x) 10^1 + a_0(x) + a_{-1}(x) 10^{\log_{10}y(-1)} + \ldots + a_{-m}(x) 10^{\log_{10}y(-m)} \nonumber 
\end{equation}

\noindent Note that $\log_{10}$ in this expression can be replaced by the natural $\log$, using $\log_{10}(x) = \frac{\log(x)}{\log(10)}$. Here, each $a_i(x) \in \{0, 1, \ldots, 9 \}$ for every index $i$ (representing the position with respect to the decimal point) and every point $x$. Thus, each function $x \rightarrow a_i(x)$ is not continuous (unless $a_i(x)$ is globally constant) and, therefore, cannot be obtained through the composition of the continuous operators in $\cal F$ (by the closure property). Note that in $\cal F$, $\log$ is the sole discontinuous operator, namely, at 0, with $\log(0) \coloneqq 0$.

\par\null\par

\noindent \textbf{Transcription factor perturbation of target gene.} Suppose $E_i$ and $E_j$ are standardized expression levels of the transcription factor $i$ and gene $j$ (say, Gaussian-distributed with mean 0 and variance 1). If $i$ is a causal regulator of $j$, then the \textit{estimated} correlation of the ``measured'' expression levels, $\widehat{\gamma_{ij}} = \textrm{corr}(\widehat{E_i}, \widehat{E_j})$, can differ substantially from the true $\gamma_{ij} = \textrm{corr}(E_i, E_j)$. The estimate $\widehat{\gamma_{ij}}$ can severely underestimate $\gamma_{ij}$. For example, in 1000 simulations $s$ (each of sample size $n = 100$) of gene expression data $\{\bm{E^s}_{i}, \bm{e^s}_{i}, \bm{E^s}_{j}, \bm{e^s}_{j}\} \in \mathbb{R}^n \times \mathbb{R}^n \times \mathbb{R}^n \times \mathbb{R}^n$, using the following ``noise'' models for $i$ and $j$, respectively:  

\begin{align}
\bm{e^s}_i = \bm{\epsilon}_1 + \bm{\epsilon}_1^{3} - \sqrt{|\bm{\epsilon}_1^{3}|} ~ \textrm{where} ~ \bm{\epsilon}_{1,1:n} \sim \mathcal{N}(0, 0.30) \nonumber \\
\bm{e^s}_j = \bm{\epsilon}_2 + \bm{\epsilon}_2^{2} + \log{|\bm{\epsilon}_2|} ~ \textrm{where} ~ \bm{\epsilon}_{2,1:n} \sim \mathcal{N}(0, 0.40) \nonumber 
\end{align}

\noindent and assuming (ground-truth) perfect correlation between $\bm{E^s}_i$ and $\bm{E^s}_j$ (in fact, $\bm{E^s}_j = \bm{E^s}_i$), the estimated correlation (average $\rho \approx 0.56$, $\mathrm{SD} \approx 0.067$) underestimates the true correlation. In addition, a non-causal relationship between $i$ and $j$ (so that $\gamma_{ij} = 0$) may show $\widehat{\gamma_{ij}} \gg 0$ due to the presence of a confounder (e.g., batch effect). Thus, the estimate $\widehat{\gamma_{ij}}$ from noisy biological data as a proxy for the perturbation effect must be interpreted with caution or supported by additional functional data. However, in empirical evaluation, $\widehat{\gamma_{ij}}$ is found to be significantly associated with $RP_{ij}(\Delta^*_i)$ by Chen et al. \cite{chen2020determinants}, with Pearson $\rho$ p-values (estimated using 2 types of omics datasets, by definition) in the range between $10^{-200}$ and $10^{-8}$.

\par\null\par

\noindent \textbf{Simulation framework to study $\psi_i$.} We conduct simulations informed by empirical data \cite{chen2020determinants}, using the observed cumulative distribution function (CDF) $F(\rho_0) \coloneqq P(r \leq \rho_0)$ of the correlation $r$ between a transcription factor, namely, the TEA domain family member 1 (TEAD1), and gene expression. Here, $F$ is invertible as it is stricly increasing. By inverse transform sampling, from the standard uniform distribution, $U \sim \textrm{Unif}[0,1]$, we simulate $X_{exp}$ that has the given CDF:  

\begin{equation}
X_{exp} \overset{d}{=} F^{-1}(U) \nonumber 
\end{equation}

\noindent Given the measured expression level $\widehat{E_i}$ of the transcription factor $i$, we simulate a target gene $j$ with measured expression level $\widehat{E_j}$ of a pre-specified correlation $r \sim P(X_{exp})$ using the dot product for vectors with mean 0:

\begin{align}
\bm{u} \cdot \bm{v} &= |\bm{u}| |\bm{v}| \cos(\theta) \nonumber \\
r &= \cos(\theta) \nonumber
\end{align}

\noindent That is, as the correlation $r$ can be geometrically interpreted as $\cos(\theta)$, we simulate a vector of angle $\theta = \arccos{(r)}$ from $\widehat{E_i}$. If $r = 1$ or $r = -1$, it is trivial to simulate $\widehat{E_j}$ (e.g., $\widehat{E_j} = \widehat{E_i}$ or $\widehat{E_j} = - \widehat{E_i}$, respectively). We assume $-1 < r < 1$. Let $\bm{v}$ be a random vector. From the orthogonal projection of $\bm{v}$ to $\bm{u}_i$, the unit vector along $\widehat{E_i}$, consider the vector $\bm{v}^\perp$ (assumed to be of length 1, without loss of generality) orthogonal to $\bm{u}_i$. Then the following vector $\bm{w}_j$ representing the simulated target gene $j$ has the desired angle and correlation with $\widehat{E_i}$:

\begin{equation} 
\bm{w}_j = \bm{v}^\perp + (\frac {\cos{(\theta)}}{\sin{(\theta)}} ) \bm{u}_i = \bm{v}^\perp + (\frac {r}{\sqrt{1-r^2}} ) \bm{u}_i \nonumber
\end{equation}

\noindent where $-1 < r < 1$. We then consider the observed values of the Regulatory Potential $RP_{ij}$ for TEAD1 \cite{chen2020determinants}. As before, we generate a random variable $X_{RP}$ that has the same CDF as observed.

\clearpage 


\clearpage

\clearpage

\clearpage

\clearpage

\clearpage

\clearpage

\clearpage

\clearpage

\clearpage

\clearpage

\clearpage

\clearpage

\end{document}